\pdfoutput=1

\documentclass[11pt]{article}

\usepackage[final]{acl}

\usepackage{times}
\usepackage{latexsym}

\usepackage[T1]{fontenc}

\usepackage[utf8]{inputenc}

\usepackage{microtype}

\usepackage{inconsolata}

\usepackage{graphicx}

\usepackage{multirow}
\usepackage{graphicx}
\usepackage{color,xcolor}
\definecolor{RED}{RGB}{247,61,65}
\definecolor{ORANGE}{RGB}{238,130,47}
\definecolor{PURPLE}{RGB}{112,48,160}
\definecolor{BLUE}{RGB}{0,112,192}
\definecolor{GREEN}{RGB}{0,176,80}
\usepackage{hyperref}
\usepackage{float}
\usepackage{subfig}
\usepackage{subcaption}
\usepackage[utf8]{inputenc}
\usepackage{appendix}
\usepackage{stfloats}
\usepackage{booktabs}
\usepackage{bbding}
\usepackage{pifont}

\title{Revealing Personality Traits: A New Benchmark Dataset for \\Explainable Personality Recognition on Dialogues}

\author{Lei Sun \textsuperscript{1}\\
  \texttt{leisun@ruc.edu.cn} \\\And
  Jinming Zhao \textsuperscript{2,$\ast$} \\
  \texttt{zhaojinming1@gmail.com} \\
  \AND
  Qin Jin \textsuperscript{1,$\ast$} \\
  \texttt{qjin@ruc.edu.cn} \\\\
  \textsuperscript{1}School of Information, Renmin University of China, 
  \textsuperscript{2}Independent Researcher
  }

\begin{document}

\maketitle
\def\thefootnote{$\dagger$}\footnotetext{Corresponding Authors.}

\begin{abstract}

Personality recognition aims to identify the personality traits implied in user data such as dialogues and social media posts.
Current research predominantly treats personality recognition as a classification task, failing to reveal the supporting evidence for the recognized personality. In this paper, we propose a novel task named \textbf{Explainable Personality Recognition}, aiming to reveal the reasoning process as supporting evidence of the personality trait.
Inspired by personality theories, personality traits are made up of stable patterns of personality state, where the states are short-term characteristic patterns of thoughts, feelings, and behaviors in a concrete situation at a specific moment in time. We propose an explainable personality recognition framework called \textbf{Chain-of-Personality-Evidence} (CoPE), which involves a reasoning process from specific contexts to short-term personality states to long-term personality traits. Furthermore, based on the CoPE framework, we construct an explainable personality recognition dataset from dialogues, \textbf{PersonalityEvd}. 
We introduce two explainable personality state recognition and explainable personality trait recognition tasks, which require models to recognize the personality state and trait labels and their corresponding support evidence.
Our extensive experiments based on Large Language Models on the two tasks show that revealing personality traits is very challenging and we present some insights for future research.
Our data and code are available at \href{https://github.com/Lei-Sun-RUC/PersonalityEvd}{https://github.com/Lei-Sun-RUC/PersonalityEvd}.


\end{abstract}
\section{Introduction}
\begin{figure}[t]
    \centering
    \includegraphics[width=0.98\linewidth]{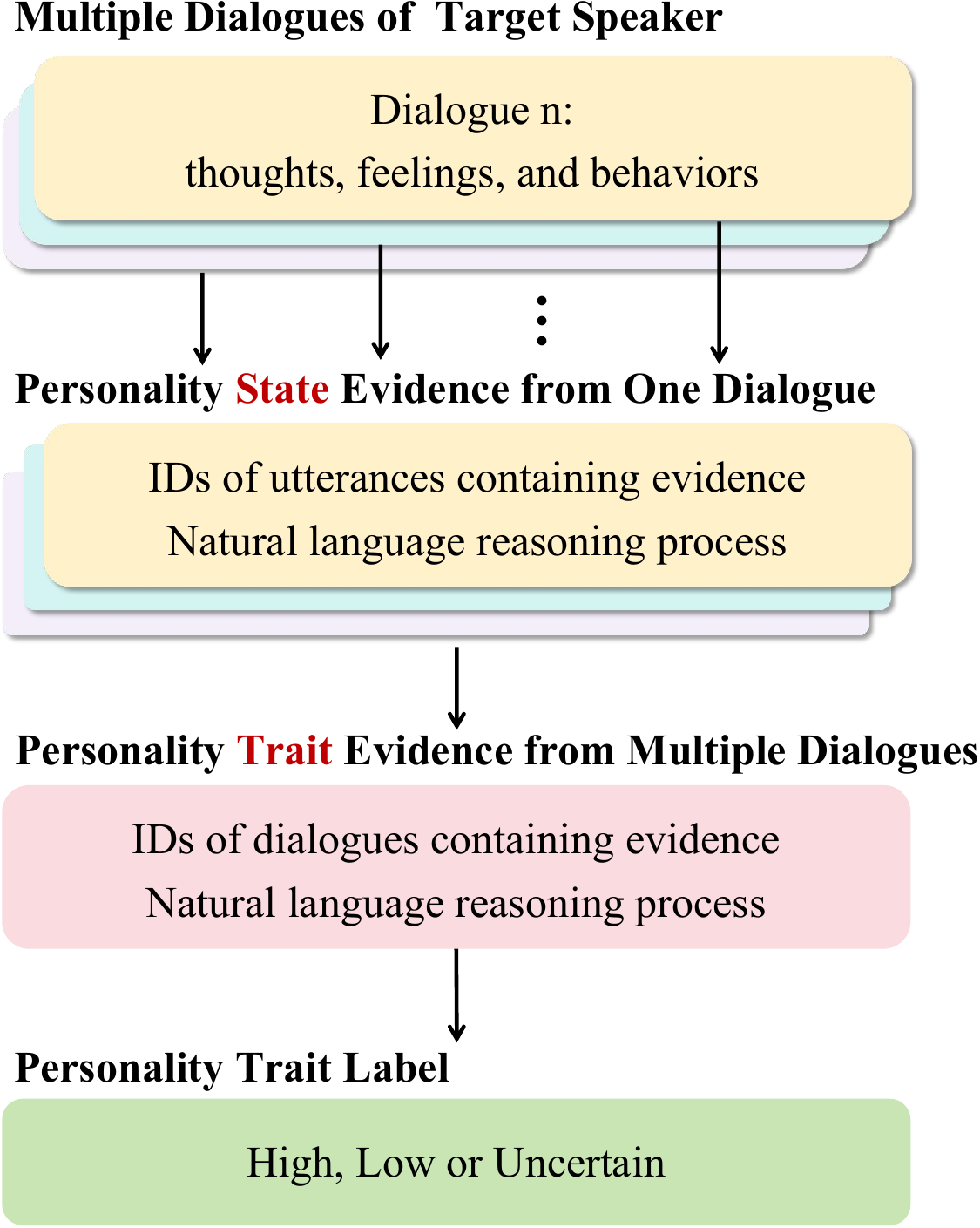}
    \caption{Chain-of-Personality-Evidence (CoPE) framework illustrates the reasoning process for revealing supporting evidence of personality traits.}
    \label{fig:COE}
\end{figure}

Personality, a characteristic way of thinking, feeling, and behaving ~\cite{roberts2009back}, has a great impact on our lives, well-being, and health. Therefore, identifying a person's personality has great potential in many real-world applications, such as human-computer interaction~\cite{attig2017assessing}, psychological diagnosis and regulation~\cite{claridge2013personality,redelmeier2021understanding}, and job candidate screenings~\cite{liem2018psychology,caldwell1998personality}.
Traditional personality recognition methods typically depend on self-reported results from designed questionnaires. Such an approach is not only time-consuming but also necessitates the cooperation of the subjects.
Therefore, Automatic Personality Recognition (APR) has attracted increasing attention in recent years, which aims to predict one's personality based on user data. To support the APR research, various personality datasets have been proposed, such as the FriendsPersona dataset~\cite{jiang2020automatic} based on dialogues, the Essays I~\cite{pennebaker1999linguistic} based on stream-of-consciousness essays, the PAN-2015 dataset~\cite{rangel2015overview} based on Twitter data, and the YouTube Vlogs dataset~\cite{biel2012youtube} based on YouTube videos.

\begin{figure*}[!t]
    \centering
    \includegraphics[width=1\linewidth]{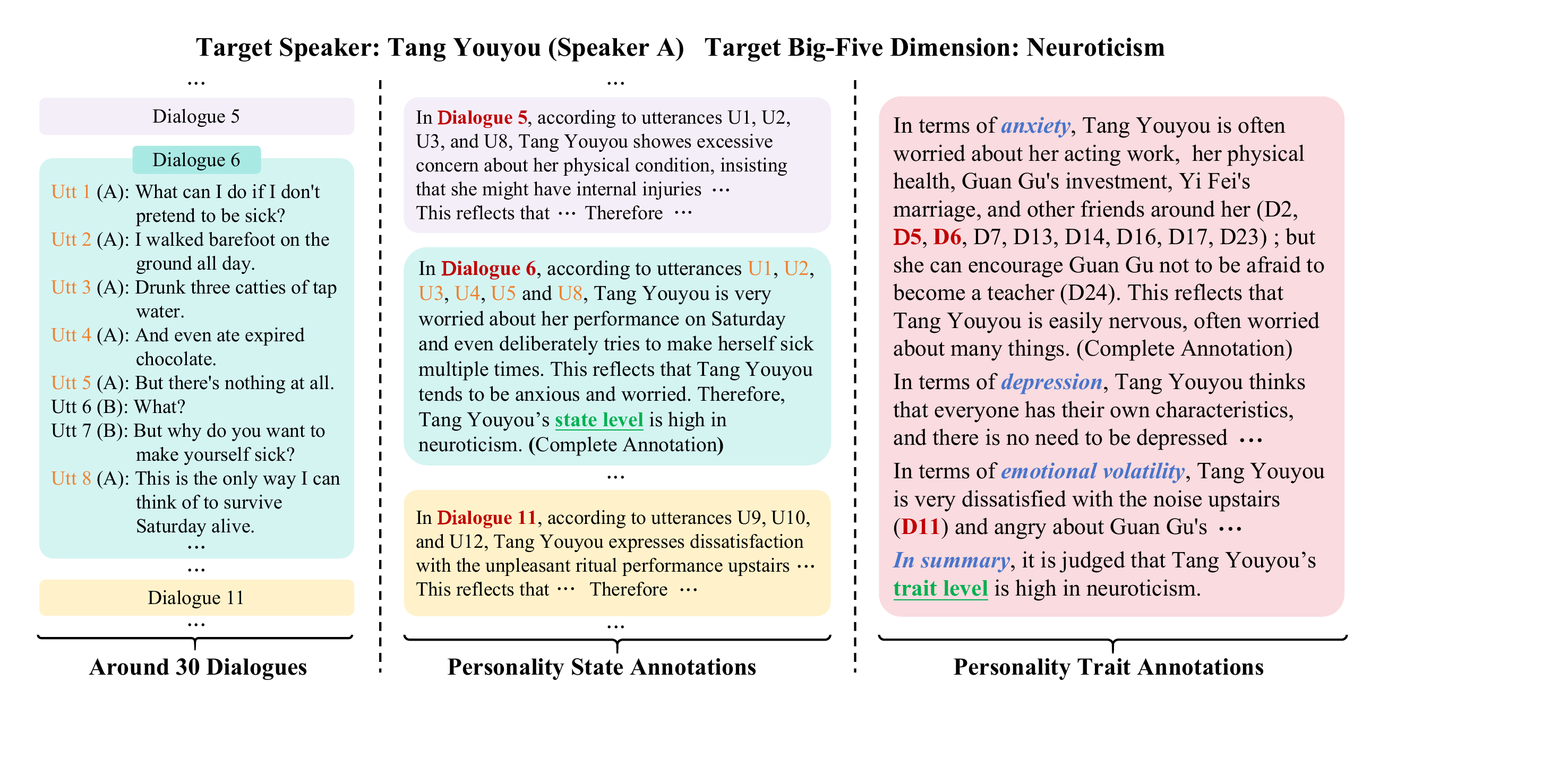}
    \caption{A Speaker's explainable personality annotation of the Neuroticism dimension from the Big-Five Personality Model, including dialogue-level personality state and speaker-level personality trait labels with corresponding evidence (natural language reasoning process). U\# denotes evidence utterances and D\# denotes evidence dialogues. The neuroticism dimension contains anxiety, depression, and emotional volatility three facets in the BFI-2 scale. For personality trait evidence annotation, we annotate the natural language reasoning process for each facet.}
    \label{fig:sample}
\end{figure*}

%
However, previous works have mainly focused on recognizing personality trait labels and fail to reveal the supporting evidence of the personality traits, making the model’s predictions uninterpreted.
Therefore, in this work, we propose a novel task to reveal personality traits named explainable personality recognition. Most theories of personality suggest that personality traits are made up of enduring patterns of personality states, where the personality states are short-term characteristic patterns of thoughts, feelings, and behaviors in a concrete situation at a specific moment in time~\cite{fleeson2001toward,fleeson2015whole}. Inspired by these theories, we propose an explanation framework called \textbf{Chain-of-Personality-Evidence} (CoPE) shown in Figure \ref{fig:COE}, which is a reasoning process to reveal the supporting evidence for personality explanation. The reasoning process goes from specific thoughts, feelings, and behaviors to short-term personality states to long-term stable personality traits. 
Due to the repetitions in one's state being critical to capture trait ~\cite{roberts2009back}, we first analyze the short-term personality state and then reveal the real stable personality trait based on all one's short-term state patterns. When revealing one's personality from dialogues, a dialogue can be seen as a short-term specific context containing thoughts, feelings, and behaviors that reflect one's personality states, and then we can obtain one's personality traits by analyzing personality states from multiple dialogues. To achieve clear and reasonable explanations, personality state evidence is comprised of evidence utterance IDs and natural language reasoning process, and trait evidence is comprised of evidence dialogue IDs and corresponding natural language reasoning process.

Furthermore, based on the proposed CoPE framework, we construct an explainable personality dataset, \textbf{PersonalityEvd}, which consists of 72 speakers and about 2000 dialogues from Chinese TV series. So each speaker is involved in around 30 dialogues. We also provide a translated English version of the dialogues. 
As shown in Figure~\ref{fig:sample}, we annotate not only dialogue-level personality state and speaker-level personality trait labels but also detailed corresponding reasoning processes to justify these labels.

Based on the proposed PersonalityEvd dataset, we propose two sub-tasks: 1) \textbf{E}vidence grounded \textbf{P}ersonality \textbf{S}tate \textbf{R}ecognition (\textbf{EPR-S}), which requires the model to predict the state label and provide prediction evidence from each dialogue. 2) \textbf{E}vidence grounded \textbf{P}ersonality \textbf{T}rait \textbf{R}ecognition (\textbf{EPR-T}), which requires the model to predict the trait label and provide prediction evidence from one's multiple dialogues. Both tasks are highly challenging, especially for the trait-level task which exists conflicts between short-term states, interactions with different interlocutors, the long context from lots of dialogues, etc. 

We further establish a strong baseline with powerful Large Language Models (LLMs) and conduct extensive experiments on the explainable personality tasks. Human evaluation results show that current LLMs are far from humans in personality understanding. 
We analyze the experimental results and find that introducing supporting evidence helps improve the performance of personality recognition. We also discover that analyzing the state evidence as an intermediate result contributes to the EPR-T task, which also proves the necessity of introducing the EPR-S task. We hope our insights can offer inspiration for further exploration.

The main contributions of this work include: 
(1) We propose a personality explanation framework called \textbf{Chain-of-Personality-Evidence} (\textbf{CoPE}) based on personality theories, which reveals a detailed reasoning process as supporting evidence of personality traits. 
(2) We manually construct a high-quality supporting dataset, \textbf{PersonalityEvd}, based on dialogues to support explainable personality recognition tasks. 
(3) We introduce two personality recognition tasks and propose a LLM-based strong baseline method. We conduct extensive experimental results and present some insights for future research.

\section{Related Work}
\subsection{Personality Theories}
Various personality theories have been developed to categorize, interpret, and understand human personality, including the Cattell Sixteen Personality Factor (16PF; ~\cite{cattell2008sixteen}), the Hans Eysenck’s psychoticism, extraversion, and neuroticism (PEN; ~\cite{eysenck2012model}), Myers-Briggs Type Indicator (MBTI; ~\cite{briggs1976myers}), the Big-Five Model~\cite{mccrae1992introduction} and so on. Among them, the most frequently used personality models are the Big-Five Model and the MBTI model. The Big-Five model measures personality through five dipolar scales: \textbf{O}penness, \textbf{C}onscientiousness, \textbf{E}xtraversion, \textbf{A}greeableness, and \textbf{N}euroticism. We utilize the Big-Five model in this work. The MBTI model lays out a binary classification based on four distinct functions (\textbf{E}xtraversion/\textbf{I}ntroversion, \textbf{S}ensing/I\textbf{N}tuition, \textbf{T}hinking/\textbf{F}eeling, \textbf{J}udgment /\textbf{P}erception). 

\subsection{Automatic Personality Recognition}
Automatic personality recognition, as an important topic in computational psycho-linguistics, focuses on determining one’s personality from a variety of data sources, such as dialogues, Twitter, Facebook, and YouTube~\cite{mushtaq2022text}. For text modality, essays are a popular mode of text and the corresponding datasets such as Essays II~\cite{tausczik2010psychological}. ~\citet{rangel2015overview} proposes the PAN-2015 corpus collected from Twitter in four languages: English, Spanish, Italian, and Dutch. The TwiSty dataset~\cite{verhoeven2016twisty} is also a multilingual twitter stylometry corpus for personality profiling. MyPersonality dataset~\cite{park2015automatic} comprises status updates of over 66,000 Facebook users. There are also several dialogue-based datasets, such as the FriendsPersona dataset~\cite{jiang2020automatic} from Friends TV Show, Story2Personality Dataset~\cite{sang-etal-2022-mbti} from movie scripts, PANDORA~\cite{gjurkovic-etal-2021-pandora} and  MBTI9k~\cite{khan2020personality} dataset sourced from Reddit posts. Some datasets with other modalities have been proposed as well, such as audio~\cite{polzehl2010automatically,mohammadi2012automatic}, visual~\cite{cristani2013unveiling}, and multimodal modalities~\cite{sanchez2013emergent,ponce2016chalearn,escalante2017design,celiktutan2017multimodal,palmero2021context}. However, no personality evidence to justify the label has been considered in all previous works. 


\section{The PersonalityEvd Dataset}
\subsection{Dialogue Selection}
We build our PersonalityEvd dataset by leveraging the CPED corpus~\cite{chen2022cped}, which is a large-scale Chinese emotional dialogue dataset containing more than 12K dialogues of 392 speakers from 40 TV shows. We first remove dialogues with less than 10 utterances because these dialogues may be incomplete or contain insufficient information. We then randomly select 72 speakers and 30 dialogues for each speaker to form a candidate set with 2,160 dialogues in total, based on which we annotate our evidence grounded dataset.

\subsection{Annotation Content}
\label{sec:annotation content}
We employ the Big Five Inventory-2 scale (BFI-2)~\cite{soto2017next,zhang2022big} as our theoretical foundation, which contains 60 short and easy-to-understand phrases. Every Big-Five dimension has 12 characteristic items, of which 6 items belong to the high level and the remaining 6 items belong to the low level. More detailed information can be found in Appendix~\ref{sec:BFI-2}.

Following previous personality recognition works ~\cite{rangel2016overview,jiang2020automatic}, we define three levels for the Big-Five personality dimensions: \textit{high, low, uncertain}. \textit{high} / \textit{low} means that the target speaker in the dialogue exhibits a high or low level of characteristics in the corresponding Big-Five dimension, while \textit{uncertain} means that the level of the target Big-Five personality dimension cannot be judged according to the dialogue. In addition to personality trait labels, we also annotate state labels, which can reflect the potential personality tendencies of the speaker.

As for the personality state evidence, it consists of three parts: \textit{evidence utterance IDs}, \textit{utterance summaries}, and \textit{personality characteristics}. \textit{Utterance summaries} and \textit{personality characteristics} together constitute natural language reasoning process of  personality states. 
For the \textit{evidence utterance IDs}, we only choose the utterances from the target speaker, although other speakers in the dialogue may provide some background information. We annotate the most relevant utterances that could reflect the personality, such as utterances within personality keywords.
The \textit{utterance summaries} are the concrete thoughts, feelings, or behaviors summarized into natural language, according to the evidence utterance IDs.
The \textit{personality characteristics} are the features that the \textit{utterance summaries} reflect, which are from the 60 personality description items in the BFI-2 psychological questionnaire. Since the 60 items of the BFI-2 are still limited and cannot cover various situations, we allow annotators to make slight changes to the items to adapt to the dialogue context while keeping the meaning consistent with the original descriptions.
Finally, we organize the three parts and the state label into the form of Chain-of-Thought (CoT)~\cite{wei2022chain,ho-etal-2023-large} using an overall description template structure, such as ``according to utterances {...(evidence utterance IDs)...}, {...(utterance summaries)...} This reflects {...(personality characteristics)...} Therefore {...(state label)...} ''. 

As for the personality trait evidence, we formulate it as a combination of three-faceted descriptions of the target Big-Five dimension. For example, the neuroticism dimension includes anxiety, depression, and emotional volatility based on the BFI-2 psychological questionnaire. Each facet description also contains three parts, and they play similar roles as state evidence components: \textit{evidence dialogue IDs}, \textit{dialogue summaries}, and \textit{personality characteristics}. \textit{Evidence dialogue IDs} is enclosed in parentheses after the corresponding \textit{dialogue summaries}.
We format them using similar templates, such as
``In terms of {(facet)}, {...(dialogue summaries)...} This reflects {...(personality characteristics)...}''. Finally, we also organize the description of these three facets and the trait label into a CoT format, as shown in Figure~\ref{fig:sample}.

\subsection{Annotation Process of Personality States}
The annotation of state evidence is highly difficult, as it requires an analysis that encompasses the five dimensions of the Big Five personality theory. To reduce annotation difficulty, improve annotation speed, and ensure high quality, we take two steps: first using GPT-4-Turbo~\cite{achiam2023gpt} to pre-annotate, and then performing manual correction.
\subsubsection{GPT-4 Pre-Annotation}
GPT-4 exhibits strong performance on various tasks, including the personality prediction and explanation task~\cite{ji2023chatgpt}. As there are five dimensions of the Big-Five Model, we handle these dimensions separately, which means that a dialogue will be taken as input five times to obtain five Big-five dimension results. Due to the limited input token length of GPT-4, we use GPT-4 to analyze five dialogues at once.  Detailed prompts are described in Appendix~\ref{sec:pre-annotation}. Because of the powerful natural language understanding capability, reasoning ability, and world knowledge, GPT-4 provides a good basis for further correction.

\subsubsection{Human Correction}
\textbf{Annotators.} Since the results of GPT-4 still contain mistakes, we screen 12 undergraduates majoring in psychology who have adequate knowledge about the Big-Five Personality Model to perform further correction. We first organize a special workshop to train them on our tasks. Each participant needs to complete trial annotations. Then feedback is provided to participants to correct their mistakes. Only when they achieve satisfactory performance on the trial dialogues can they start annotating the main dataset.

\begin{figure*}[!t]
\centering
\subfloat[]
{\includegraphics[width=0.33\linewidth]{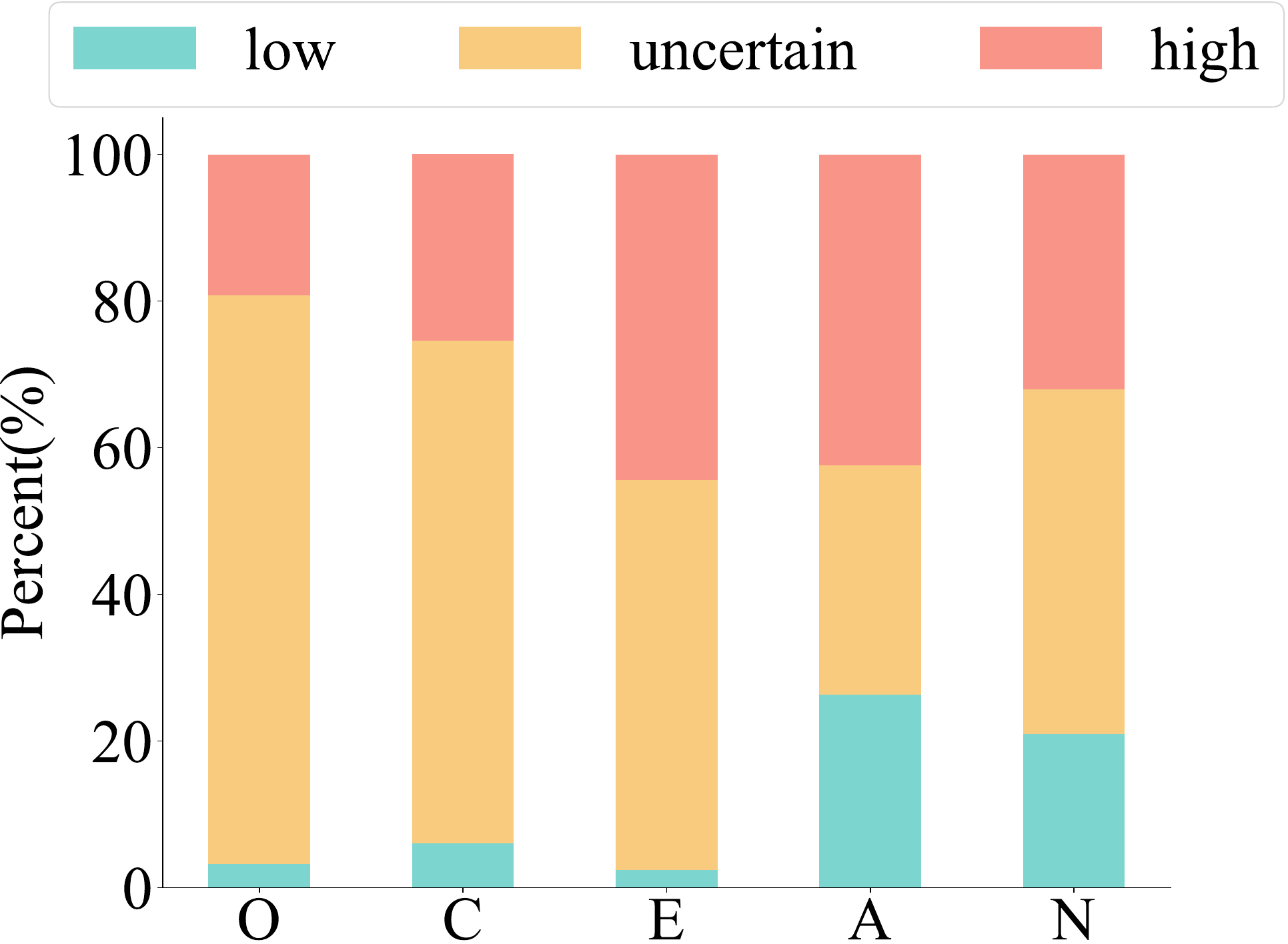}}\hspace{1pt} 
\subfloat[]
{\includegraphics[width=0.33\linewidth]{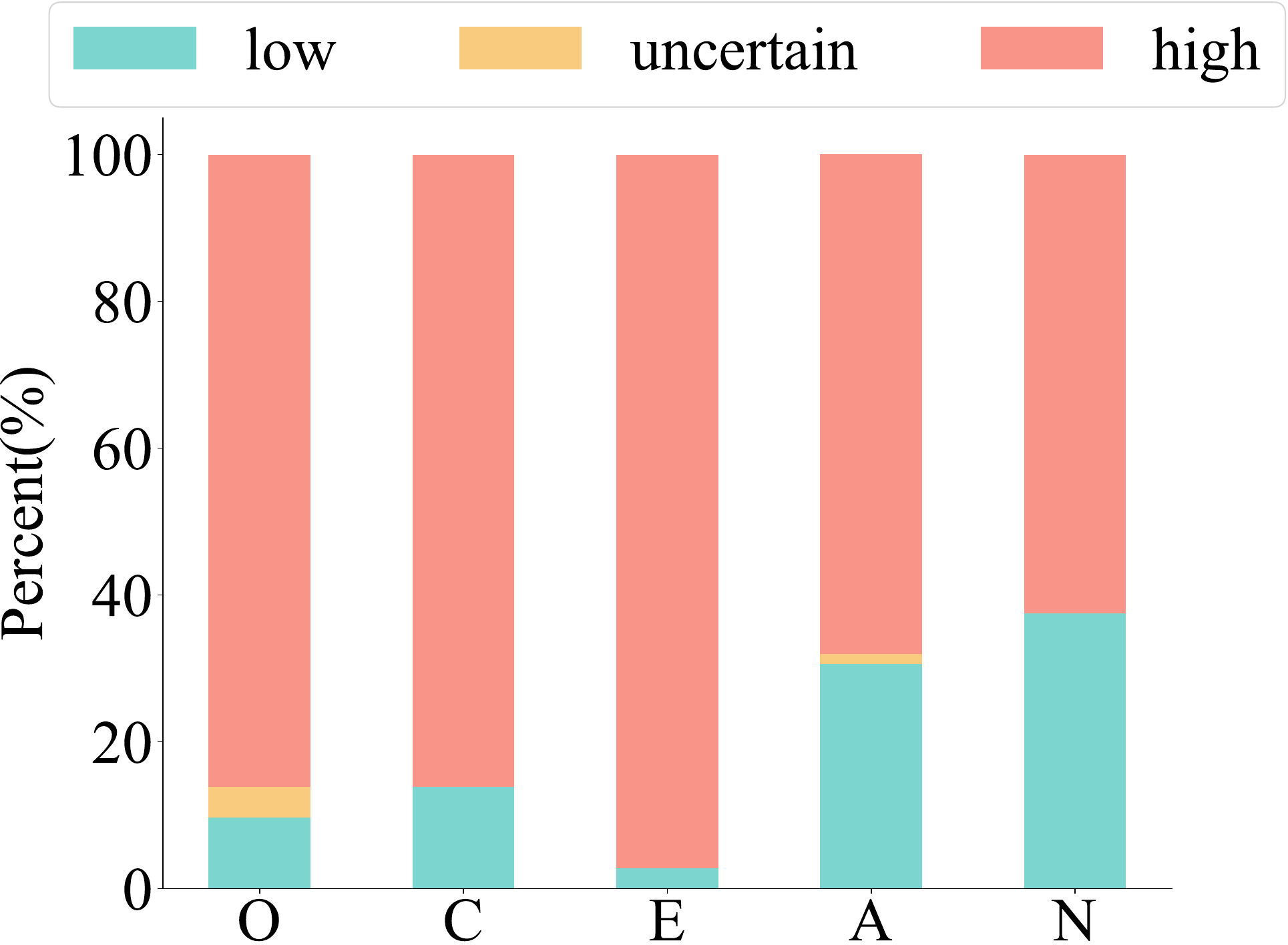}} 
\subfloat[]
{\includegraphics[width=0.3\linewidth]{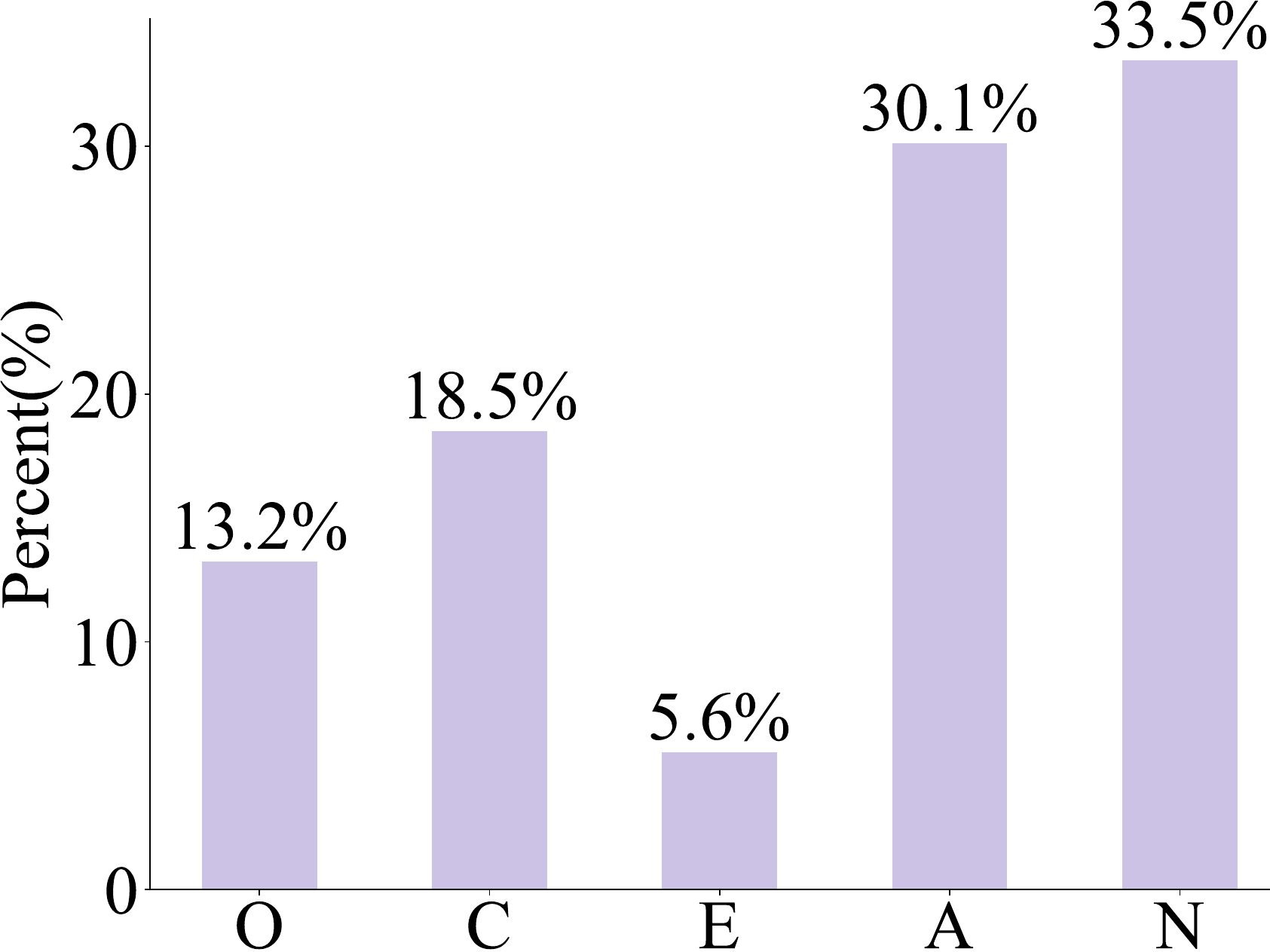}}
\caption{(a) The distribution of state labels. (b) The distribution of trait labels. (c) The ratio of the state label different from the trait label.  (O: openness, C: conscientiousness, E: extraversion,
A: agreeableness, N: neuroticism)}
\label{fig:personality_label}
\end{figure*}

\noindent \textbf{Annotation Guidelines.} Each sample is first annotated by one annotator. The annotator first understands the dialogue context and then makes corrections based on the results predicted by GPT-4. We ask annotators to analyze obvious personalities and not to over-interpret or speculate on the dialogue. For evidence utterance IDs, we ask the annotator to annotate the most relevant sentence IDs; for utterance summaries and personality characteristics, if the analysis of GPT-4 is reasonable and comprehensive, the annotator needs to simplify the content to highlight the key points. If the analysis is wrong, the annotator needs to re-annotate. Approximately 30\% of the samples were re-annotated manually, greatly improving the efficiency of our dataset construction.

There are two new quality inspectors who are graduate students with more specialized knowledge in personality theory, ensuring higher accuracy in the annotations. They give the annotator suggestions for modifications to help the annotator refine the annotation. If there is a disagreement between the quality inspector and the annotator, they will discuss to reach a consensus. We remove samples where there is no consensus. The two quality inspectors review all the samples to ensure the annotation quality.

\noindent \textbf{Final Check.} At the final stage, we, the authors, manually review, filter, and correct all the annotations again. We design several strategies to filter and correct the unsatisfactory annotations: 1) We remove dialogues whose state labels of the five Big-Five personality dimensions are all \textit{uncertain} labels, as these dialogues cannot provide evidence information and thus don't match the focus of our dataset. The number of these dialogues is small and does not affect our dataset scale. 2) We remove dialogues with contradictory personality information. Although it is reasonable that one shows contradictory personalities, this will make it difficult to annotate the state label. 3) We correct the language errors, such as typos, misnomers, etc.

\subsection{Annotation Process of Personality Traits}
\textbf{Annotators.}
After completing the personality state annotations, we recruit another 6 undergraduates majoring in psychology to finish the personality trait annotations. We train them using similar training steps as the state labeling for trait annotation.

\noindent \textbf{Annotation Guidelines.} Each sample is first labeled individually by 3 annotators. Annotators are asked to score and write the evidence according to the BFI-2 scale based on about 30 dialogues and previous state annotations. To maintain the consistency of state and trait annotations, annotators are allowed to modify the state annotation if they feel that the original annotation is unreasonable. Some dialogues are difficult to understand because the dialogue scenes are difficult to infer, involve many characters, etc., and different annotators may interpret the dialogue differently.

After getting individual annotations, the 3 annotators will reach a consensus through discussion, including the score and corresponding evidence. To obtain the trait label, we calculate the average score of 12 BFI-2 items corresponding to each dimension and convert it to binary labels using the median split. 

\noindent \textbf{Final Check.} 
In the end, we manually review and correct all trait annotations again: 1) we update the state annotations modified in this stage into final results. 2) We also check for language errors.

\begin{table}[t]
    \small
    \centering
    \scalebox{1.2}{
    \begin{tabular}{l|c} \toprule
        \textbf{Statistics} & \textbf{Number} \\ \midrule
        \# Dialogues & 1,924 \\
        \# Speakers & 72 \\
        \# Utterances & 32,673 \\
        Avg. dlg per spk & 26.72 \\
        Avg. utt per dlg & 16.98 \\
        Avg. length of utt  & 16.73 \\
        Avg. tgt spk utt per dlg & 8.43 \\
        Avg. evi utt IDs per dlg & 4.26 \\
        Avg. evi dlg IDs per spk & 11.96 \\
        SD. dlg per spk & 2.47 \\
        SD. utt per dlg & 5.36 \\
        SD. length of utt & 4.76 \\
        SD. tgt spk utt per dlg & 4.01 \\
        SD. evi utt IDs per dlg & 2.87 \\
        SD. evi dlg IDs per spk & 5.88 \\

        \bottomrule
    \end{tabular}
    }
    \caption{\textbf{PersonalityEvd} dataset statistics. (utt, tgt, spk, dlg, evi, SD refer to  utterance, target, speaker, dialogue, evidence, standard deviation.)
    }
    \label{tab:dia_sta}
\end{table}

\begin{figure*}[t]
\centering
\subfloat[High Openness]{\includegraphics[width=0.22\linewidth]{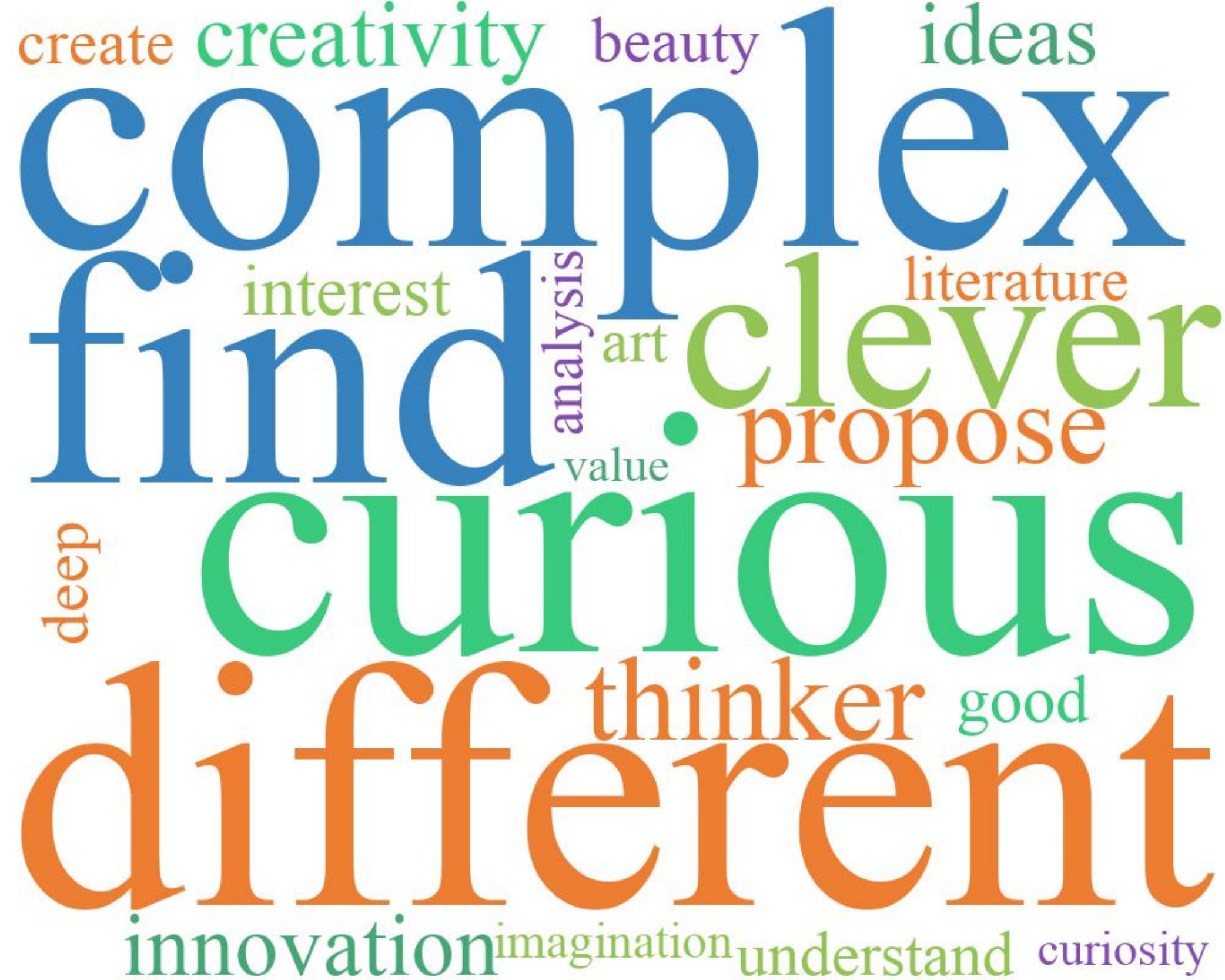}} \hspace{10pt}
\subfloat[Low Openness]{\includegraphics[width=0.22\linewidth]{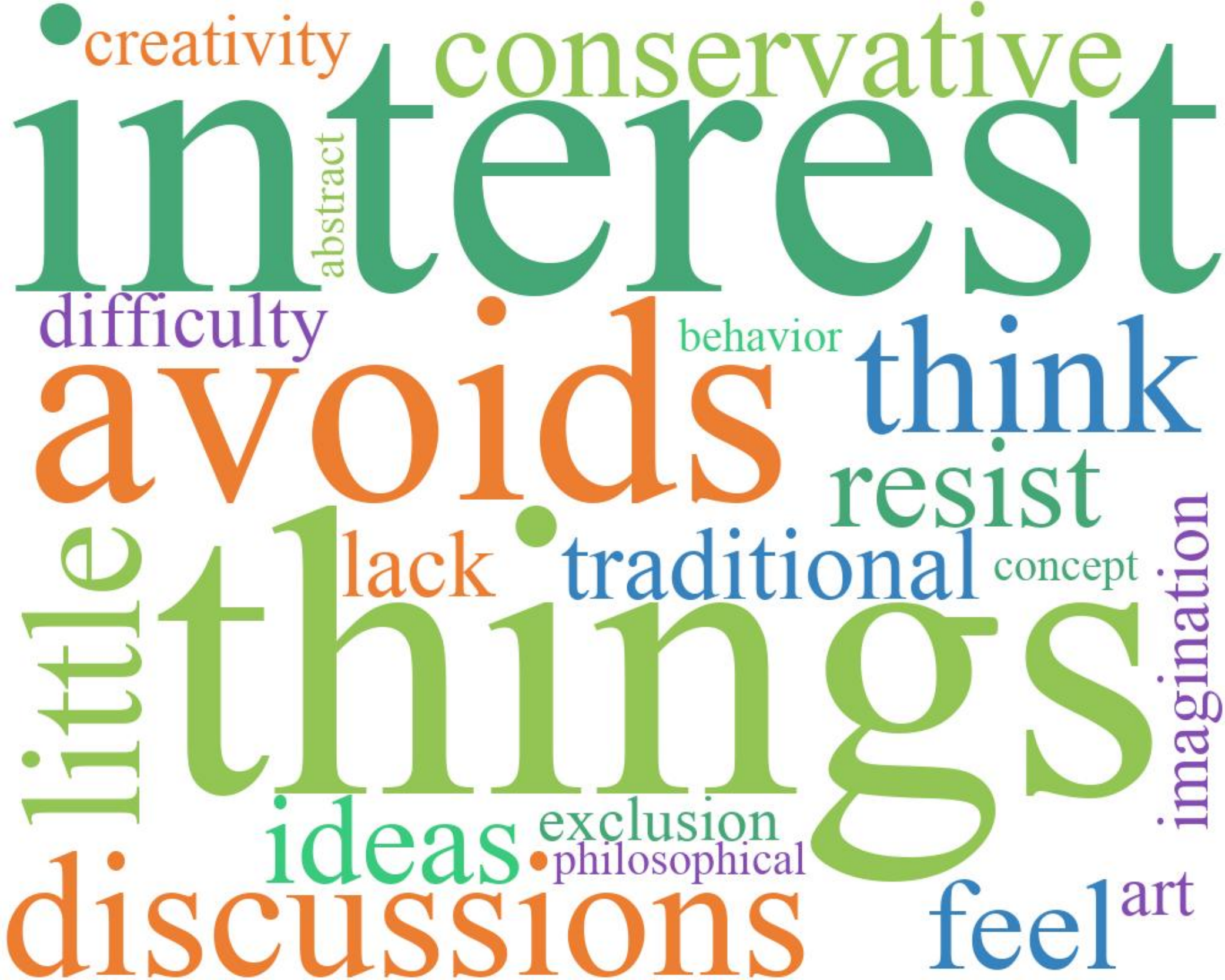}} \hspace{10pt}
\subfloat[High Conscientiousness]{\includegraphics[width=0.22\linewidth]{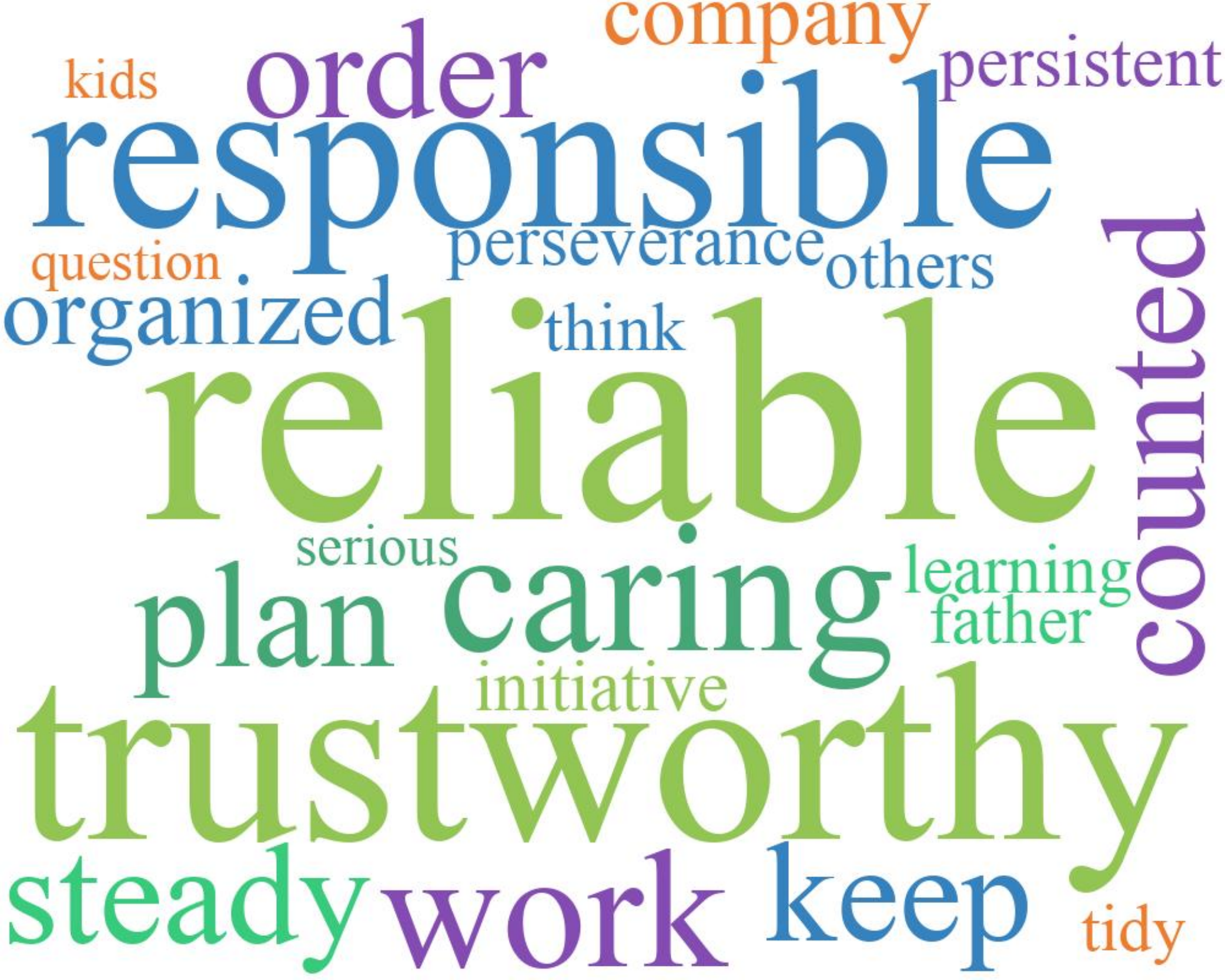}} \hspace{10pt}
\subfloat[Low Conscientiousness]{\includegraphics[width=0.22\linewidth]{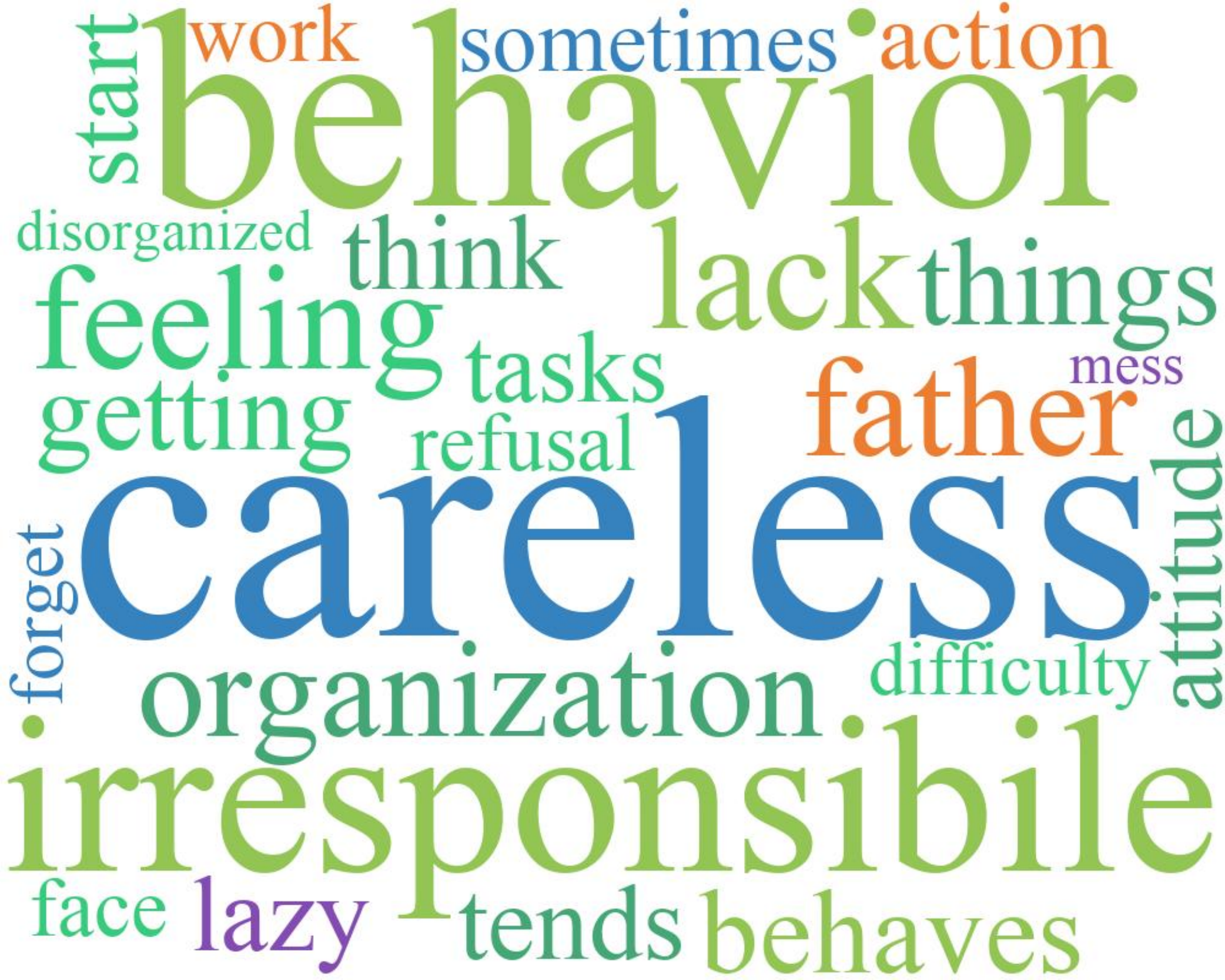}}
\caption{The word clouds of personality \textbf{states} reasoning process on openness and conscientiousness dimensions.}
\label{fig:wordcloud}
\end{figure*}

\subsection{Dataset Statistics}

\noindent \textbf{Dialogue Statistics.}
Table \ref{tab:dia_sta} presents the statistics of our constructed PersonalityEvd dataset. It contains 1,924 dialogues of 72 speakers, which means that every speaker is involved in approximately 30 dialogues. The average number of utterances per dialogue is 16.98, indicating long context and rich information in our dataset. The average number of utterances and dialogue IDs containing evidence is 4.26 and 11.96, respectively. The average number of target speaker utterances percentage per dialogue is about 50\%, indicating the balanced involvement in dialogues.

\noindent \textbf{Annotation Statistics.}
Figure \ref{fig:personality_label} presents the distribution of two-level personality labels and the ratio of the state label when it's different from the trait label. As for the state labels, we can see that \textit{uncertain} labels occupy a relatively large proportion due to the sparsity of personality information in dialogue. As for the trait labels, \textit{high} labels account for a significant proportion as characters of TV series often have distinct personalities. Figure (c) shows the ratio of the state label when it’s different from the trait label, which illustrates the daily variance of personality.

Figure \ref{fig:wordcloud} presents the top frequently used words in the natural language reasoning process of different state labels of the openness and conscientiousness dimensions, which can reflect the personality keywords of our PersonalityEvd dataset.
For example, in the openness dimension, words like ``different'' and ``curious''  show a high level of openness. The word clouds of remaining three dimensions are shown in Appendix~\ref{sec:left would cloud}.

\noindent \textbf{Dataset splits.}
For the state-level data, we randomly split it into train/valid/test sets based on speakers according to the ratio of 7:1:2, ensuring that the speakers in the training set do not appear in the valid or test set. 
For the limited trait-level data, we randomly divide the data into 3 folds based on speakers, so each fold contains 24 speakers.

\section{Proposed Tasks}
We introduce two novel sub-tasks based on our dataset: {E}vidence grounded {P}ersonality {S}tate {R}ecognition (\textbf{EPR-S}) and {E}vidence grounded {P}ersonality {T}rait {R}ecognition (\textbf{EPR-T}).

\subsection{The EPR-S Task}
\textbf{Definition} There are five Big-Five personality dimensions \(BF=[bf_1, bf_2, ..., bf_5 ]\). Each dialogue \(D\) includes \(m\) speakers \(P= [p_1, p_2, ..., p_m] (m\geq2)\). The EPR-S task aims to recognize the state label \(y_s\in \{high, low, uncertain\}\) as well as generate its evidence \(E_s\) of the target speaker \(p_i\) for the target Big-Five personality dimension \(bf_j\) given one dialogue \(D\).

\noindent \textbf{Evaluation}
As the prediction of the EPR-S task contains four parts, we evaluate them separately. For the evaluation of evidence utterance IDs, we report the binary F1-score. For the evaluation of the natural language reasoning process, consisting of utterance summaries and personality characteristics, we use BERTScore~\cite{zhang2019bertscore} to measure the semantic similarity between the ground truth and predicted evidence. As BERTScore is still a limited metric, we also use claude-3-sonnet-20240229~\cite{anthropic2024claude} and gpt-4-turbo-2024-04-09~\cite{achiam2023gpt} to evaluate the semantic overlapping level from 1 to 5~\cite{li2024leveraging}. For the evaluation of the personality label, to be consistent with previous works~\cite{majumder2017deep,jiang2020automatic,guo2024personality}, we use accuracy as the evaluation metric.

\subsection{The EPR-T Task}
\textbf{Definition} The EPR-T task aims to  recognize the trait label \(y_t\in \{high, low, uncertain\}\) as well as generate corresponding evidence \(E_t\) of the target speaker \(p\) for the target Big-Five personality dimension \(bf_i\) given dialogues \([D_1, D_2, ..., D_n]\).

\noindent \textbf{Evaluation}
The prediction of the EPR-T task has a similar structure as the EPR-S task. For the evaluation of evidence dialogue IDs and trait labels, we apply the same metrics as the EPR-S task. For the natural language trait evidence, we compute the average similarity score of the three facets between the ground truth and predicted description, using the same three models mentioned above.
\section{Experiments}

\begin{table*}[ht]
\centering
\small
\renewcommand\arraystretch{1.1}
\scalebox{1.08}{
\begin{tabular}{cc|c|ccc|cccccc}
\toprule[1.2pt]
\multirow{2}{*}{Task} & \multirow{2}{*}{Model} & ID F1 & BERT & Claude & GPT-4 & \multicolumn{6}{c}{Personality Accuracy} \\ \cmidrule{3-12} 
&   & Avg  & Avg & Avg   & Avg & O   & C   & E   & A   & N   & Avg  \\ \midrule
\multirow{3}{*}{EPR-S} & GLM-32k & 75.42 & \textbf{83.45} & 3.61  & 3.48 & 74.10  & 68.59 & 63.36 & 55.64 & 61.43 & 64.62 \\
& Qwen-32k  & \textbf{75.94} & 83.44 &  \textbf{3.68} & \textbf{3.56}  & 75.48 & \textbf{68.87} & \textbf{63.39} & 63.36 & 61.15 & \textbf{66.45} \\
& GPT-4  &  71.20 & 76.21  &  3.55  & 3.27 & \textbf{76.85} & 50.41 & 50.96 & \textbf{69.97} & \textbf{62.25} & 62.09  \\ \midrule[1.2pt]
\multirow{2}{*}{EPR-T} & GLM-32k   & 40.28 &  76.80  &  2.95  & 2.74  & \textbf{81.76} & \textbf{86.11} & 95.77  & \textbf{73.12} & 52.17 & \textbf{77.78} \\
& Qwen-32k  & \textbf{44.39} & \textbf{77.81} &  \textbf{3.25}  & \textbf{3.11}  & 74.75 & 75.96  & \textbf{97.16} & 70.41 & \textbf{64.67} & 76.59 \\ \bottomrule[1.2pt]
\end{tabular}
}
\caption{Model performance of CoT fine-tuning on two tasks. ID F1 denotes the binary F1-score of evidence utterance/dialogue IDs. BERT, Claude, and GPT-4 refer to BERTScore (F1), Claude-3-sonnet, and GPT-4-Turbo score. The Claude-3-sonnet and GPT-4-Turbo scores range from 1 to 5. CoT fine-tuning: the model is trained to generate the evidence and then the answer, as shown in Figure \ref{fig:sample}. For the EPR-T task, we report the results of 3-fold cross-validation. (O: openness, C: conscientiousness, E: extraversion, A: agreeableness, N: neuroticism)}
 \label{tab:mian_results}
\end{table*}


\subsection{Baselines}

We evaluate three baseline LLMs on our tasks:

\noindent $\bullet$ \textbf{ChatGLM3-6B-32K} based on ChatGLM3-6B~\cite{du-etal-2022-glm,zeng2022glm}, further strengthens the ability to understand long texts and can better handle contexts up to 32K in length.
 
\noindent $\bullet$ \textbf{Qwen1.5-7B-Chat} is the improved version of Qwen~\cite{bai2023qwen}, which has significant model quality improvements in chat models and supports 32K context length.
 
\noindent  $\bullet$ \textbf{GPT-4-Turbo-2024-04-09} is a snapshot of GPT-4-Turbo~\cite{achiam2023gpt} from April 9th, 2024 with more powerful performance and lower price than GPT-4.

We denote them as \textbf{GLM-32k}, \textbf{Qwen-32k}, and \textbf{GPT-4} respectively in the following sub-sections.

\begin{table}[t]
\centering
\renewcommand\arraystretch{1.1}
\small
\scalebox{1.05}{
\begin{tabular}{c|ccc}
\toprule[1.2pt]
Model & Fluency & Coherence & Plausibility  \\ \midrule
Ground Truth      & \textbf{4.61} & \textbf{4.38} &  \textbf{4.31}      \\
GLM-32k    & 3.82 & 2.51 & 2.59    \\
Qwen-32k   & 3.90& 2.68  & 2.65     \\ \bottomrule[1.2pt]
\end{tabular}
}

\caption{Results of human evaluation on the natural language reasoning process of personality \textbf{traits}. The score range is 1 to 5.}
 \label{tab:human evaluation}
\end{table}

\subsection{Main Results}
Tabel \ref{tab:mian_results} presents the results of two tasks on the test set. We evaluate GPT-4-Turbo in the zero-shot setting and other models in the LoRA~\cite{hu2021lora} fine-tuning (FT) setting.
For the EPR-S task, Qwen1.5-7B-Chat achieves the best overall performance among the three models. Surprisingly, GPT-4-Turbo achieves comparable results in the zero-shot setting and even surpasses the other two models in openness, agreeableness, and neuroticism dimensions in terms of personality accuracy.
For the EPR-T task, in terms of average personality accuracy, ChatGLM3-6B-32K outperforms Qwen1.5-7B-Chat, while Qwen1.5-7B-Chat surpasses ChatGLM3-6B-32K on evidence-related metrics. In summary, the scores of these models on both tasks are still low, implying significant room for further improvements.

To avoid potential bias in the GPT-4-Turbo evaluation of self-generated content, we also use Claude-3-sonnet for evaluation. In all experiments, Claude-3-sonnet scores slightly higher than GPT-4-Turbo, indicating that GPT-4-Turbo does not give higher scores to its own generated results.

\subsection{Human Evaluation}

We evaluate the natural language reasoning process of the traits, concerning the following criteria: 

\noindent    $\bullet$ \textbf{Fluency} measures the grammatical and formatting aspects of the sentences.
    
\noindent    $\bullet$ \textbf{Coherence} measures whether the text is semantically and factually consistent with the dialogue context.
    
\noindent    $\bullet$ \textbf{Plausibility} measures whether the text contains comprehensive and correct trait evidence.

We randomly selected 50 samples of 10 speakers for human evaluation. The annotation was performed by undergraduates majoring in psychology, and we assigned five evaluators to each sample. They assessed each aspect on a scale from 1 to 5, with higher scores indicating better results. Finally, we calculate the average scores of these evaluators. Results are reported in Table~\ref{tab:human evaluation}. The three scores of ground truth are very close to 5 points, so we can conclude that the quality of PersonalityEvd is guaranteed. The fluency scores of ChatGLM3-6B-32K and Qwen1.5-7B-Chat reach 3.82 and 3.90 respectively, probably because current LLMs are trained on a large corpus and have mastered human language. Both models have low coherence and plausibility scores, demonstrating the great challenge of explaining personality.

\subsection{Ablation Study}
\textbf{Potential benefits of corresponding evidence for personality recognition.}
In addition to CoT fine-tuning, we provide the results of Hybrid fine-tuning, inspired by the method in ~\cite{li2024explanations}, as shown in Table \ref{tab:evidence_ablation}. We use the ChatGLM3-6B-32k model for the validation. For state recognition, the performance of  Hybrid-Direct setting and Hybrid-CoT is 2.1\% and 1.71\% higher than that of Direct fine-tuning, respectively, proving that introducing evidence helps improve the reasoning ability of LLMs. The performance of CoT fine-tuning is declined compared to the result of Direct fine-tuning. We speculate that it is because predicting evidence increases the task difficulty, even if the CoT helps improve the model's reasoning ability.

However, for trait recognition, the performance of both Hybrid settings is similar to that of Direct settings. We speculate that the model has not been fully trained due to the limited training data. In the CoT setting, the performance has improved, implying that  improving the model's reasoning ability can overcome the increase in task difficulty.

\begin{table}[t]
\centering
\small
\scalebox{1.1}{
\begin{tabular}{c|cc|c}
\toprule[1.2pt]
Task                   & Fine-tuning & Inference & Avg accuracy \\ \midrule
\multirow{4}{*}{State} & Direct & Direct & 64.84    \\
                       & CoT & CoT & 64.62  \\ 
                       & Hybrid & Direct & \textbf{66.94}  \\
                       & Hybrid & CoT & 66.55 \\ \midrule[1.2pt]
\multirow{4}{*}{Trait} & Direct & Direct & 75.83 \\
                       & CoT & CoT &  \textbf{77.78}  \\ 
                       & Hybrid & Direct &  75.53  \\ 
                       & Hybrid & CoT & 75.55 \\\bottomrule[1.2pt] 
\end{tabular}
}
\caption{Ablation study on GLM-32k model to prove the benefits of introducing evidence on personality recognition. Direct: the model is trained or evaluated to directly generate the answer; CoT: the model is trained or evaluated to generate the evidence and then the answer; Hybrid: the model is trained on both above tasks.}
\label{tab:evidence_ablation}
\end{table}

\begin{table}[t]
\centering
\renewcommand\arraystretch{1.1}
\small
\begin{tabular}{c|c|ccc|c}
\toprule[1.2pt] 
State & ID F1 & BERT & Claude & GPT-4 & Accuracy \\ \midrule
 None     & 40.28 & 76.80 & 2.95  & 2.74 & 77.78 \\
 Pred     & 44.18 & 76.99  & 3.31  & 3.11  & 77.99  \\
GT      & \textbf{77.09} & \textbf{81.03} &  \textbf{3.69} & \textbf{3.57} & \textbf{83.52} \\ \bottomrule[1.2pt] 
\end{tabular}
\caption{Average personality trait metrics when different state clues act as inputs on GLM-32k to prove the necessity of introducing the EPR-S task. None: no state clues, just dialogues; Pred: predicted state evidence from GLM-32k; GT: ground-truth state evidence.}
 \label{tab:state4trait}
\end{table}

\noindent \textbf{Potential benefits of the state evidence for the EPR-T task.}
Table \ref{tab:state4trait} presents the results when different state clues act as the input, which can prove the necessity of introducing the EPR-S task.
We can find that when providing the state evidence predicted by the model itself, all metrics are slightly improved. Using the state evidence as input instead of the original dialogues can avoid handling very long contexts and reduce the difficulty of the task. However, because the current model's ability to predict state evidence is poor, the improvement is very limited.
When providing ground-truth state evidence, all metrics are significantly improved, as gold references rarely contain noise information. Therefore, for the trait-level task, we believe that adopting a two-stage pipeline method is a promising direction that can be attempted in the future. 

\section{Conclusion}
In this paper, to promote the research of explainable personality recognition, we propose a novel Chain-of-Personality-Evidence (CoPE) framework, which reveals the reasoning process from specific contexts to short-term personality states to long-term personality traits. We build an explainable personality dataset based on CoPE framework, namely PersonalityEvd, which supports two explainable personality recognition tasks (EPR-S and EPR-T) that both require the model to recognize personality as well as provide corresponding evidence.
Finally, we evaluate several large language models as baselines and conduct extensive experiments on both tasks. Additional human evaluations validate the quality of our constructed dataset and our analytical experiments present insights for future work. We hope that our PersonalityEvd dataset and two novel tasks can facilitate further investigation into explainable personality analysis in the community.

\section*{Limitations}
There are several limitations of our PersonalityEvd dataset and modeling. First, our dataset is in Chinese. Though we translated the dataset from Chinese into English, we think it is better to directly construct a corresponding English dataset considering data quality. Second, we acknowledge that our dataset is small-scale due to the high annotation costs of two-level personality evidence. However, we plan to expand the dataset scale in the future. Third, as our dataset is constructed around characters, standardized and psychometrically validated personality tests or self-report questionnaires could not be applied to provide objective evidence for the explanations. Last, we just essentially fine-tune the LLMs as baselines, since we are more focused on the contributions of the dataset construction and building the benchmark. We leave the development of advanced models as future work.

\section*{Ethics Statements}
We choose the dialogues from the CPED dataset, which has been released to the public. There are no intellectual property disputes for our data source. Human annotation is carried out by workers we employ and we pay each worker \$12 per hour, which is a fair and reasonable hourly wage in Beijing. Besides, due to the subjectivity of manual annotation, our dataset may contain biased opinions.
\section*{Acknowledgements}
We thank all reviewers for their insightful comments and suggestions.
This work was supported by  the National Natural Science Foundation of China (No. 62072462).


\bibliography{custom}

\begin{thebibliography}{49}
\providecommand{\natexlab}[1]{#1}

\bibitem[{Achiam et~al.(2023)Achiam, Adler, Agarwal, Ahmad, Akkaya, Aleman, Almeida, Altenschmidt, Altman, Anadkat et~al.}]{achiam2023gpt}
Josh Achiam, Steven Adler, Sandhini Agarwal, Lama Ahmad, Ilge Akkaya, Florencia~Leoni Aleman, Diogo Almeida, Janko Altenschmidt, Sam Altman, Shyamal Anadkat, et~al. 2023.
\newblock Gpt-4 technical report.
\newblock \emph{arXiv preprint arXiv:2303.08774}.

\bibitem[{Anthropic(2024)}]{anthropic2024claude}
AI~Anthropic. 2024.
\newblock The claude 3 model family: Opus, sonnet, haiku.
\newblock \emph{Claude-3 Model Card}.

\bibitem[{Attig et~al.(2017)Attig, Wessel, and Franke}]{attig2017assessing}
Christiane Attig, Daniel Wessel, and Thomas Franke. 2017.
\newblock Assessing personality differences in human-technology interaction: an overview of key self-report scales to predict successful interaction.
\newblock In \emph{HCI International 2017--Posters' Extended Abstracts: 19th International Conference, HCI International 2017, Vancouver, BC, Canada, July 9--14, 2017, Proceedings, Part I 19}, pages 19--29. Springer.

\bibitem[{Bai et~al.(2023)Bai, Bai, Chu, Cui, Dang, Deng, Fan, Ge, Han, Huang et~al.}]{bai2023qwen}
Jinze Bai, Shuai Bai, Yunfei Chu, Zeyu Cui, Kai Dang, Xiaodong Deng, Yang Fan, Wenbin Ge, Yu~Han, Fei Huang, et~al. 2023.
\newblock Qwen technical report.
\newblock \emph{arXiv preprint arXiv:2309.16609}.

\bibitem[{Biel and Gatica-Perez(2012)}]{biel2012youtube}
Joan-Isaac Biel and Daniel Gatica-Perez. 2012.
\newblock The youtube lens: Crowdsourced personality impressions and audiovisual analysis of vlogs.
\newblock \emph{IEEE Transactions on Multimedia}, 15(1):41--55.

\bibitem[{Briggs(1976)}]{briggs1976myers}
Katharine~C Briggs. 1976.
\newblock \emph{Myers-Briggs type indicator}.
\newblock Consulting Psychologists Press Palo Alto, CA.

\bibitem[{Caldwell and Burger(1998)}]{caldwell1998personality}
David~F Caldwell and Jerry~M Burger. 1998.
\newblock Personality characteristics of job applicants and success in screening interviews.
\newblock \emph{Personnel Psychology}, 51(1):119--136.

\bibitem[{Cattell and Mead(2008)}]{cattell2008sixteen}
Heather~EP Cattell and Alan~D Mead. 2008.
\newblock The sixteen personality factor questionnaire (16pf).
\newblock \emph{The SAGE handbook of personality theory and assessment}, 2:135--159.

\bibitem[{Celiktutan et~al.(2017)Celiktutan, Skordos, and Gunes}]{celiktutan2017multimodal}
Oya Celiktutan, Efstratios Skordos, and Hatice Gunes. 2017.
\newblock Multimodal human-human-robot interactions (mhhri) dataset for studying personality and engagement.
\newblock \emph{IEEE Transactions on Affective Computing}, 10(4):484--497.

\bibitem[{Chen et~al.(2022)Chen, Fan, Xing, Pang, Huang, Han, Tie, and Xu}]{chen2022cped}
Yirong Chen, Weiquan Fan, Xiaofen Xing, Jianxin Pang, Minlie Huang, Wenjing Han, Qianfeng Tie, and Xiangmin Xu. 2022.
\newblock Cped: A large-scale chinese personalized and emotional dialogue dataset for conversational ai.
\newblock \emph{arXiv preprint arXiv:2205.14727}.

\bibitem[{Claridge and Davis(2013)}]{claridge2013personality}
Gordon Claridge and Caroline Davis. 2013.
\newblock \emph{Personality and psychological disorders}.
\newblock Routledge.

\bibitem[{Cristani et~al.(2013)Cristani, Vinciarelli, Segalin, and Perina}]{cristani2013unveiling}
Marco Cristani, Alessandro Vinciarelli, Cristina Segalin, and Alessandro Perina. 2013.
\newblock Unveiling the multimedia unconscious: Implicit cognitive processes and multimedia content analysis.
\newblock In \emph{Proceedings of the 21st ACM international conference on Multimedia}, pages 213--222.

\bibitem[{Du et~al.(2022)Du, Qian, Liu, Ding, Qiu, Yang, and Tang}]{du-etal-2022-glm}
Zhengxiao Du, Yujie Qian, Xiao Liu, Ming Ding, Jiezhong Qiu, Zhilin Yang, and Jie Tang. 2022.
\newblock \href {https://doi.org/10.18653/v1/2022.acl-long.26} {{GLM}: General language model pretraining with autoregressive blank infilling}.
\newblock In \emph{Proceedings of the 60th Annual Meeting of the Association for Computational Linguistics (Volume 1: Long Papers)}, pages 320--335, Dublin, Ireland. Association for Computational Linguistics.

\bibitem[{Escalante et~al.(2017)Escalante, Guyon, Escalera, Jacques, Madadi, Bar{\'o}, Ayache, Viegas, G{\"u}{\c{c}}l{\"u}t{\"u}rk, G{\"u}{\c{c}}l{\"u} et~al.}]{escalante2017design}
Hugo~Jair Escalante, Isabelle Guyon, Sergio Escalera, Julio Jacques, Meysam Madadi, Xavier Bar{\'o}, Stephane Ayache, Evelyne Viegas, Ya{\u{g}}mur G{\"u}{\c{c}}l{\"u}t{\"u}rk, Umut G{\"u}{\c{c}}l{\"u}, et~al. 2017.
\newblock Design of an explainable machine learning challenge for video interviews.
\newblock In \emph{2017 International joint conference on neural networks (IJCNN)}, pages 3688--3695. IEEE.

\bibitem[{Eysenck(2012)}]{eysenck2012model}
Hans~Jurgen Eysenck. 2012.
\newblock \emph{A model for personality}.
\newblock Springer Science \& Business Media.

\bibitem[{Fleeson(2001)}]{fleeson2001toward}
William Fleeson. 2001.
\newblock Toward a structure-and process-integrated view of personality: Traits as density distributions of states.
\newblock \emph{Journal of personality and social psychology}, 80(6):1011.

\bibitem[{Fleeson and Jayawickreme(2015)}]{fleeson2015whole}
William Fleeson and Eranda Jayawickreme. 2015.
\newblock Whole trait theory.
\newblock \emph{Journal of research in personality}, 56:82--92.

\bibitem[{Gjurkovi{\'c} et~al.(2021)Gjurkovi{\'c}, Karan, Vukojevi{\'c}, Bo{\v{s}}njak, and Snajder}]{gjurkovic-etal-2021-pandora}
Matej Gjurkovi{\'c}, Mladen Karan, Iva Vukojevi{\'c}, Mihaela Bo{\v{s}}njak, and Jan Snajder. 2021.
\newblock \href {https://doi.org/10.18653/v1/2021.socialnlp-1.12} {{PANDORA} talks: Personality and demographics on {R}eddit}.
\newblock In \emph{Proceedings of the Ninth International Workshop on Natural Language Processing for Social Media}, pages 138--152, Online. Association for Computational Linguistics.

\bibitem[{Guo et~al.(2024)Guo, Hirai, Ohashi, Chiba, Tsunomori, and Higashinaka}]{guo2024personality}
Ao~Guo, Ryu Hirai, Atsumoto Ohashi, Yuya Chiba, Yuiko Tsunomori, and Ryuichiro Higashinaka. 2024.
\newblock Personality prediction from task-oriented and open-domain human--machine dialogues.
\newblock \emph{Scientific Reports}, 14(1):3868.

\bibitem[{Ho et~al.(2023)Ho, Schmid, and Yun}]{ho-etal-2023-large}
Namgyu Ho, Laura Schmid, and Se-Young Yun. 2023.
\newblock \href {https://doi.org/10.18653/v1/2023.acl-long.830} {Large language models are reasoning teachers}.
\newblock In \emph{Proceedings of the 61st Annual Meeting of the Association for Computational Linguistics (Volume 1: Long Papers)}, pages 14852--14882, Toronto, Canada. Association for Computational Linguistics.

\bibitem[{Hu et~al.(2021)Hu, Shen, Wallis, Allen-Zhu, Li, Wang, Wang, and Chen}]{hu2021lora}
Edward~J Hu, Yelong Shen, Phillip Wallis, Zeyuan Allen-Zhu, Yuanzhi Li, Shean Wang, Lu~Wang, and Weizhu Chen. 2021.
\newblock Lora: Low-rank adaptation of large language models.
\newblock \emph{arXiv preprint arXiv:2106.09685}.

\bibitem[{Ji et~al.(2023)Ji, Wu, Zheng, Hu, Chen, and He}]{ji2023chatgpt}
Yu~Ji, Wen Wu, Hong Zheng, Yi~Hu, Xi~Chen, and Liang He. 2023.
\newblock Is chatgpt a good personality recognizer? a preliminary study.
\newblock \emph{arXiv preprint arXiv:2307.03952}.

\bibitem[{Jiang et~al.(2020)Jiang, Zhang, and Choi}]{jiang2020automatic}
Hang Jiang, Xianzhe Zhang, and Jinho~D Choi. 2020.
\newblock Automatic text-based personality recognition on monologues and multiparty dialogues using attentive networks and contextual embeddings (student abstract).
\newblock In \emph{Proceedings of the AAAI conference on artificial intelligence}, volume~34, pages 13821--13822.

\bibitem[{Khan et~al.(2020)Khan, Ahmad, Asghar, Saddozai, Arif, and Khalid}]{khan2020personality}
Alam~Sher Khan, Hussain Ahmad, Muhammad~Zubair Asghar, Furqan~Khan Saddozai, Areeba Arif, and Hassan~Ali Khalid. 2020.
\newblock Personality classification from online text using machine learning approach.
\newblock \emph{International journal of advanced computer science and applications}, 11(3):460--476.

\bibitem[{Li et~al.(2024{\natexlab{a}})Li, Chen, yelong shen, Chen, Zhang, Li, Wang, Qian, Peng, Mao, Chen, and Yan}]{li2024explanations}
Shiyang Li, Jianshu Chen, yelong shen, Zhiyu Chen, Xinlu Zhang, Zekun Li, Hong Wang, Jing Qian, Baolin Peng, Yi~Mao, Wenhu Chen, and Xifeng Yan. 2024{\natexlab{a}}.
\newblock \href {https://openreview.net/forum?id=rH8ZUcfL9r} {Explanations from large language models make small reasoners better}.
\newblock In \emph{2nd Workshop on Sustainable AI}.

\bibitem[{Li et~al.(2024{\natexlab{b}})Li, Xu, Shen, Xu, Gu, and Tao}]{li2024leveraging}
Zhen Li, Xiaohan Xu, Tao Shen, Can Xu, Jia-Chen Gu, and Chongyang Tao. 2024{\natexlab{b}}.
\newblock Leveraging large language models for nlg evaluation: A survey.
\newblock \emph{arXiv preprint arXiv:2401.07103}.

\bibitem[{Liem et~al.(2018)Liem, Langer, Demetriou, Hiemstra, Sukma~Wicaksana, Born, and K{\"o}nig}]{liem2018psychology}
Cynthia~CS Liem, Markus Langer, Andrew Demetriou, Annemarie~MF Hiemstra, Achmadnoer Sukma~Wicaksana, Marise~Ph Born, and Cornelius~J K{\"o}nig. 2018.
\newblock Psychology meets machine learning: Interdisciplinary perspectives on algorithmic job candidate screening.
\newblock \emph{Explainable and interpretable models in computer vision and machine learning}, pages 197--253.

\bibitem[{Majumder et~al.(2017)Majumder, Poria, Gelbukh, and Cambria}]{majumder2017deep}
Navonil Majumder, Soujanya Poria, Alexander Gelbukh, and Erik Cambria. 2017.
\newblock \href {https://doi.org/10.1109/MIS.2017.23} {Deep learning-based document modeling for personality detection from text}.
\newblock \emph{IEEE Intelligent Systems}, 32(2):74--79.

\bibitem[{McCrae and John(1992)}]{mccrae1992introduction}
Robert~R McCrae and Oliver~P John. 1992.
\newblock An introduction to the five-factor model and its applications.
\newblock \emph{Journal of personality}, 60(2):175--215.

\bibitem[{Mohammadi and Vinciarelli(2012)}]{mohammadi2012automatic}
Gelareh Mohammadi and Alessandro Vinciarelli. 2012.
\newblock Automatic personality perception: Prediction of trait attribution based on prosodic features.
\newblock \emph{IEEE Transactions on Affective Computing}, 3(3):273--284.

\bibitem[{Mushtaq and Kumar(2022)}]{mushtaq2022text}
Sumiya Mushtaq and Neerendra Kumar. 2022.
\newblock Text-based automatic personality recognition: Recent developments.
\newblock In \emph{Proceedings of Third International Conference on Computing, Communications, and Cyber-Security: IC4S 2021}, pages 537--549. Springer.

\bibitem[{Palmero et~al.(2021)Palmero, Selva, Smeureanu, Junior, Jacques, Clap{\'e}s, Mosegu{\'\i}, Zhang, Gallardo, Guilera et~al.}]{palmero2021context}
Cristina Palmero, Javier Selva, Sorina Smeureanu, Julio Junior, CS~Jacques, Albert Clap{\'e}s, Alexa Mosegu{\'\i}, Zejian Zhang, David Gallardo, Georgina Guilera, et~al. 2021.
\newblock Context-aware personality inference in dyadic scenarios: Introducing the udiva dataset.
\newblock In \emph{Proceedings of the IEEE/CVF Winter Conference on Applications of Computer Vision}, pages 1--12.

\bibitem[{Park et~al.(2015)Park, Schwartz, Eichstaedt, Kern, Kosinski, Stillwell, Ungar, and Seligman}]{park2015automatic}
Gregory Park, H~Andrew Schwartz, Johannes~C Eichstaedt, Margaret~L Kern, Michal Kosinski, David~J Stillwell, Lyle~H Ungar, and Martin~EP Seligman. 2015.
\newblock Automatic personality assessment through social media language.
\newblock \emph{Journal of personality and social psychology}, 108(6):934.

\bibitem[{Pennebaker and King(1999)}]{pennebaker1999linguistic}
James~W Pennebaker and Laura~A King. 1999.
\newblock Linguistic styles: language use as an individual difference.
\newblock \emph{Journal of personality and social psychology}, 77(6):1296.

\bibitem[{Polzehl et~al.(2010)Polzehl, M{\"o}ller, and Metze}]{polzehl2010automatically}
Tim Polzehl, Sebastian M{\"o}ller, and Florian Metze. 2010.
\newblock Automatically assessing personality from speech.
\newblock In \emph{2010 IEEE fourth international conference on semantic computing}, pages 134--140. IEEE.

\bibitem[{Ponce-L{\'o}pez et~al.(2016)Ponce-L{\'o}pez, Chen, Oliu, Corneanu, Clap{\'e}s, Guyon, Bar{\'o}, Escalante, and Escalera}]{ponce2016chalearn}
V{\'\i}ctor Ponce-L{\'o}pez, Baiyu Chen, Marc Oliu, Ciprian Corneanu, Albert Clap{\'e}s, Isabelle Guyon, Xavier Bar{\'o}, Hugo~Jair Escalante, and Sergio Escalera. 2016.
\newblock Chalearn lap 2016: First round challenge on first impressions-dataset and results.
\newblock In \emph{Computer Vision--ECCV 2016 Workshops: Amsterdam, The Netherlands, October 8-10 and 15-16, 2016, Proceedings, Part III 14}, pages 400--418. Springer.

\bibitem[{Rangel et~al.(2015)Rangel, Celli, Rosso, Potthast, Stein, Daelemans et~al.}]{rangel2015overview}
Francisco Rangel, Fabio Celli, Paolo Rosso, Martin Potthast, Benno Stein, Walter Daelemans, et~al. 2015.
\newblock Overview of the 3rd author profiling task at pan 2015.
\newblock In \emph{CLEF2015 Working Notes. Working Notes of CLEF 2015-Conference and Labs of the Evaluation forum.} Notebook Papers.

\bibitem[{Rangel et~al.(2016)Rangel, Rosso, Verhoeven, Daelemans, Potthast, and Stein}]{rangel2016overview}
Francisco Rangel, Paolo Rosso, Ben Verhoeven, Walter Daelemans, Martin Potthast, and Benno Stein. 2016.
\newblock Overview of the 4th author profiling task at pan 2016: cross-genre evaluations.
\newblock In \emph{Working Notes Papers of the CLEF 2016 Evaluation Labs. CEUR Workshop Proceedings/Balog, Krisztian [edit.]; et al.}, pages 750--784.

\bibitem[{Redelmeier et~al.(2021)Redelmeier, Najeeb, and Etchells}]{redelmeier2021understanding}
Donald~A Redelmeier, Umberin Najeeb, and Edward~E Etchells. 2021.
\newblock Understanding patient personality in medical care: five-factor model.
\newblock \emph{Journal of General Internal Medicine}, 36:2111--2114.

\bibitem[{Roberts(2009)}]{roberts2009back}
Brent~W Roberts. 2009.
\newblock Back to the future: Personality and assessment and personality development.
\newblock \emph{Journal of research in personality}, 43(2):137--145.

\bibitem[{Sanchez-Cortes et~al.(2013)Sanchez-Cortes, Aran, Jayagopi, Schmid~Mast, and Gatica-Perez}]{sanchez2013emergent}
Dairazalia Sanchez-Cortes, Oya Aran, Dinesh~Babu Jayagopi, Marianne Schmid~Mast, and Daniel Gatica-Perez. 2013.
\newblock Emergent leaders through looking and speaking: from audio-visual data to multimodal recognition.
\newblock \emph{Journal on Multimodal User Interfaces}, 7:39--53.

\bibitem[{Sang et~al.(2022)Sang, Mou, Yu, Wang, Li, and Stanton}]{sang-etal-2022-mbti}
Yisi Sang, Xiangyang Mou, Mo~Yu, Dakuo Wang, Jing Li, and Jeffrey Stanton. 2022.
\newblock \href {https://doi.org/10.18653/v1/2022.findings-emnlp.500} {{MBTI} personality prediction for fictional characters using movie scripts}.
\newblock In \emph{Findings of the Association for Computational Linguistics: EMNLP 2022}, pages 6715--6724, Abu Dhabi, United Arab Emirates. Association for Computational Linguistics.

\bibitem[{Soto and John(2017)}]{soto2017next}
Christopher~J Soto and Oliver~P John. 2017.
\newblock The next big five inventory (bfi-2): Developing and assessing a hierarchical model with 15 facets to enhance bandwidth, fidelity, and predictive power.
\newblock \emph{Journal of personality and social psychology}, 113(1):117.

\bibitem[{Tausczik and Pennebaker(2010)}]{tausczik2010psychological}
Yla~R Tausczik and James~W Pennebaker. 2010.
\newblock The psychological meaning of words: Liwc and computerized text analysis methods.
\newblock \emph{Journal of language and social psychology}, 29(1):24--54.

\bibitem[{Verhoeven et~al.(2016)Verhoeven, Daelemans, and Plank}]{verhoeven2016twisty}
Ben Verhoeven, Walter Daelemans, and Barbara Plank. 2016.
\newblock Twisty: a multilingual twitter stylometry corpus for gender and personality profiling.
\newblock In \emph{Proceedings of the Tenth international conference on language resources and evaluation (LREC'16)}, pages 1632--1637.

\bibitem[{Wei et~al.(2022)Wei, Wang, Schuurmans, Bosma, Xia, Chi, Le, Zhou et~al.}]{wei2022chain}
Jason Wei, Xuezhi Wang, Dale Schuurmans, Maarten Bosma, Fei Xia, Ed~Chi, Quoc~V Le, Denny Zhou, et~al. 2022.
\newblock Chain-of-thought prompting elicits reasoning in large language models.
\newblock \emph{Advances in neural information processing systems}, 35:24824--24837.

\bibitem[{Zeng et~al.(2022)Zeng, Liu, Du, Wang, Lai, Ding, Yang, Xu, Zheng, Xia et~al.}]{zeng2022glm}
Aohan Zeng, Xiao Liu, Zhengxiao Du, Zihan Wang, Hanyu Lai, Ming Ding, Zhuoyi Yang, Yifan Xu, Wendi Zheng, Xiao Xia, et~al. 2022.
\newblock Glm-130b: An open bilingual pre-trained model.
\newblock \emph{arXiv preprint arXiv:2210.02414}.

\bibitem[{Zhang et~al.(2022)Zhang, Li, Li, Luo, Ye, Yin, Chen, Soto, and John}]{zhang2022big}
Bo~Zhang, Yi~Ming Li, Jian Li, Jing Luo, Yonghao Ye, Lu~Yin, Zhuosheng Chen, Christopher~J Soto, and Oliver~P John. 2022.
\newblock The big five inventory--2 in china: A comprehensive psychometric evaluation in four diverse samples.
\newblock \emph{Assessment}, 29(6):1262--1284.

\bibitem[{Zhang et~al.(2019)Zhang, Kishore, Wu, Weinberger, and Artzi}]{zhang2019bertscore}
Tianyi Zhang, Varsha Kishore, Felix Wu, Kilian~Q Weinberger, and Yoav Artzi. 2019.
\newblock Bertscore: Evaluating text generation with bert.
\newblock \emph{arXiv preprint arXiv:1904.09675}.

\end{thebibliography}
\clearpage
\appendix
\section*{Appendix}

\section{Facets and Items of The BFI-2 Scale}
\label{sec:BFI-2}
\begin{figure*}[ht]
    \centering
    \includegraphics[width=1\linewidth]{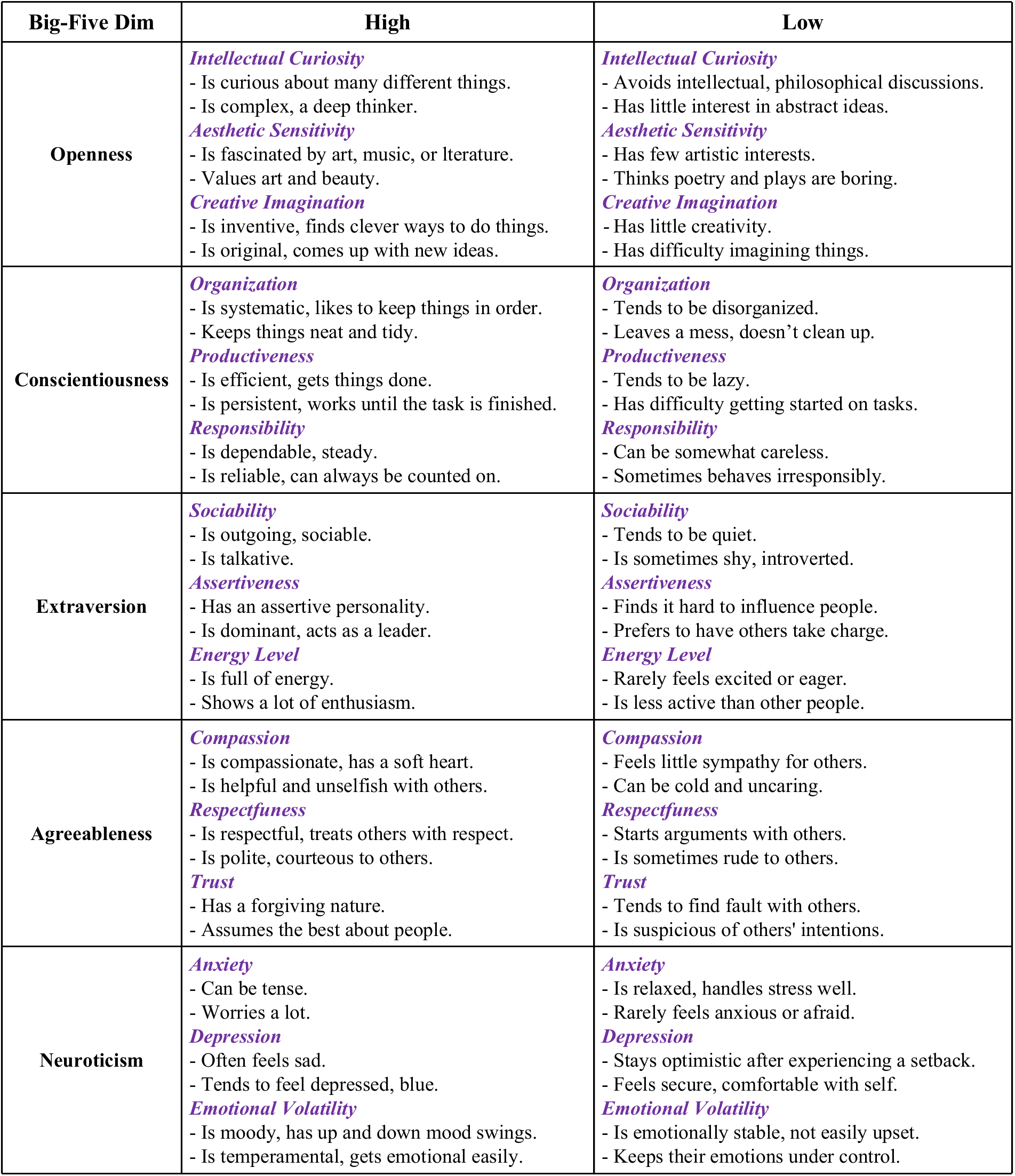}
    \caption{\textcolor{PURPLE}{Facets} and Items of The BFI-2 Scale. (Dim: dimension)}
    \label{fig:BFI-2}
\end{figure*}

The description of \textit{personality statistics} source from the BFI-2 Scale~\cite{soto2017next}. Figure \ref{fig:BFI-2} presents the facets and items of each Big-Five dimensions.

\section{Prompt for GPT-4 Pre-annotation}
\label{sec:pre-annotation}
\begin{figure*}[ht]
    \centering
    \includegraphics[width=1\linewidth]{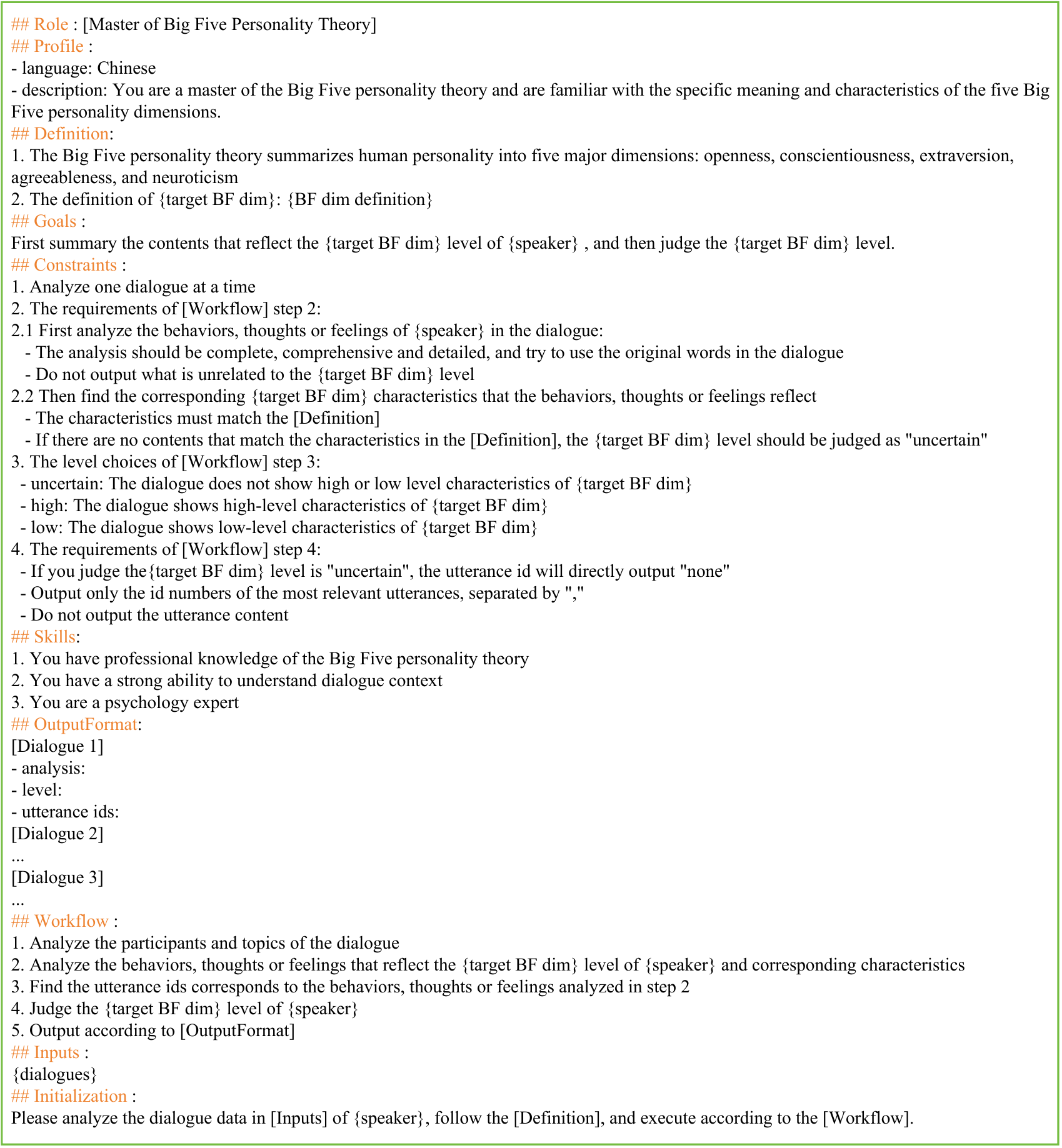}
    \caption{The prompt for GPT-4 pre-annotation. (BF: Big-Five, dim: dimension)}
    \label{fig:GPT-4-Pre-Annotation}
\end{figure*}

As shown in Figure \ref{fig:GPT-4-Pre-Annotation}, we design the prompt for GPT-4 pre-annotation.

\section{Word Clouds of Left Dimensions}
\label{sec:left would cloud}
\begin{figure*}[ht]
\centering
\subfloat[High Ext.]{\includegraphics[width=0.15\linewidth]{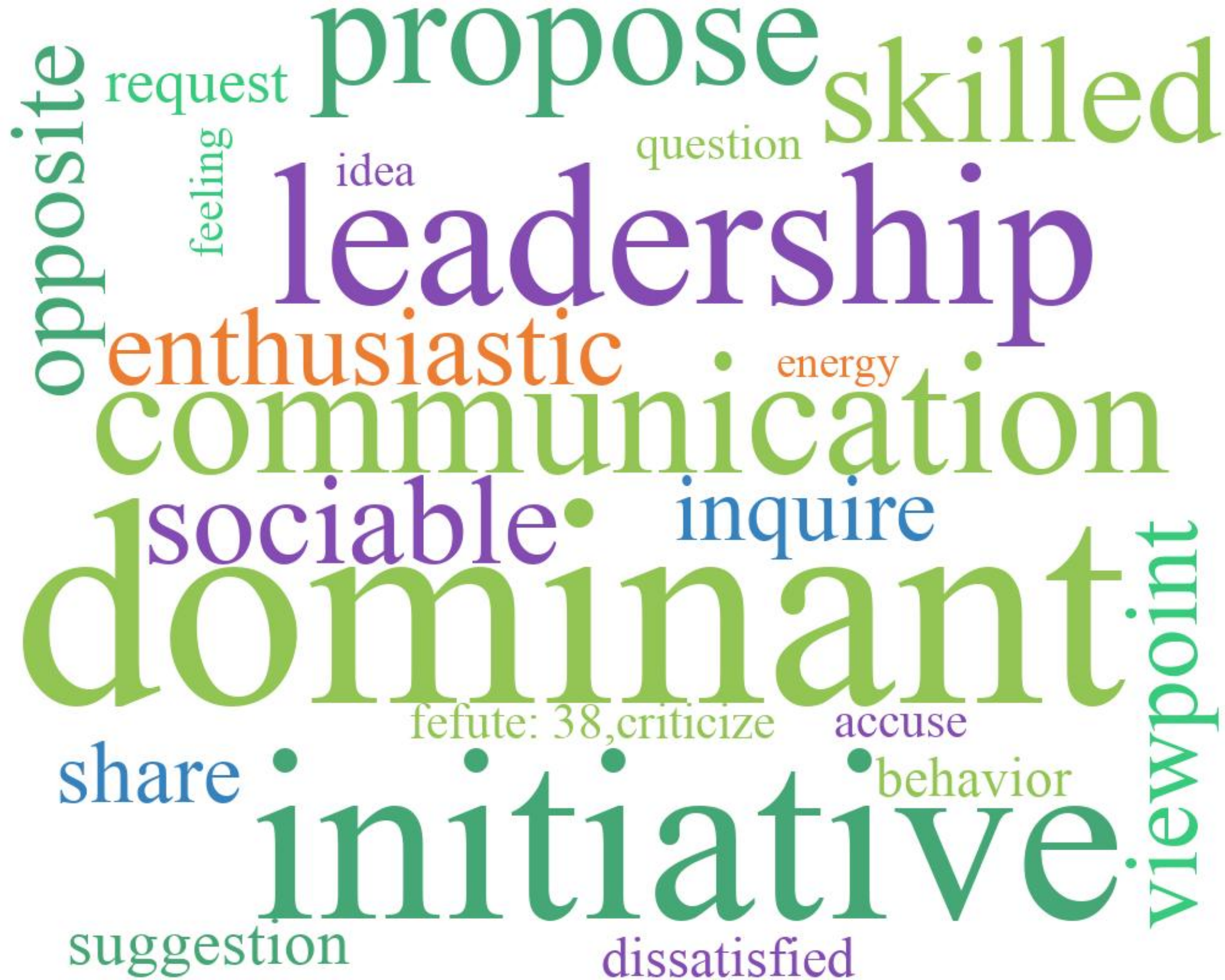}} \hspace{5pt}
\subfloat[Low Ext.]{\includegraphics[width=0.15\linewidth]{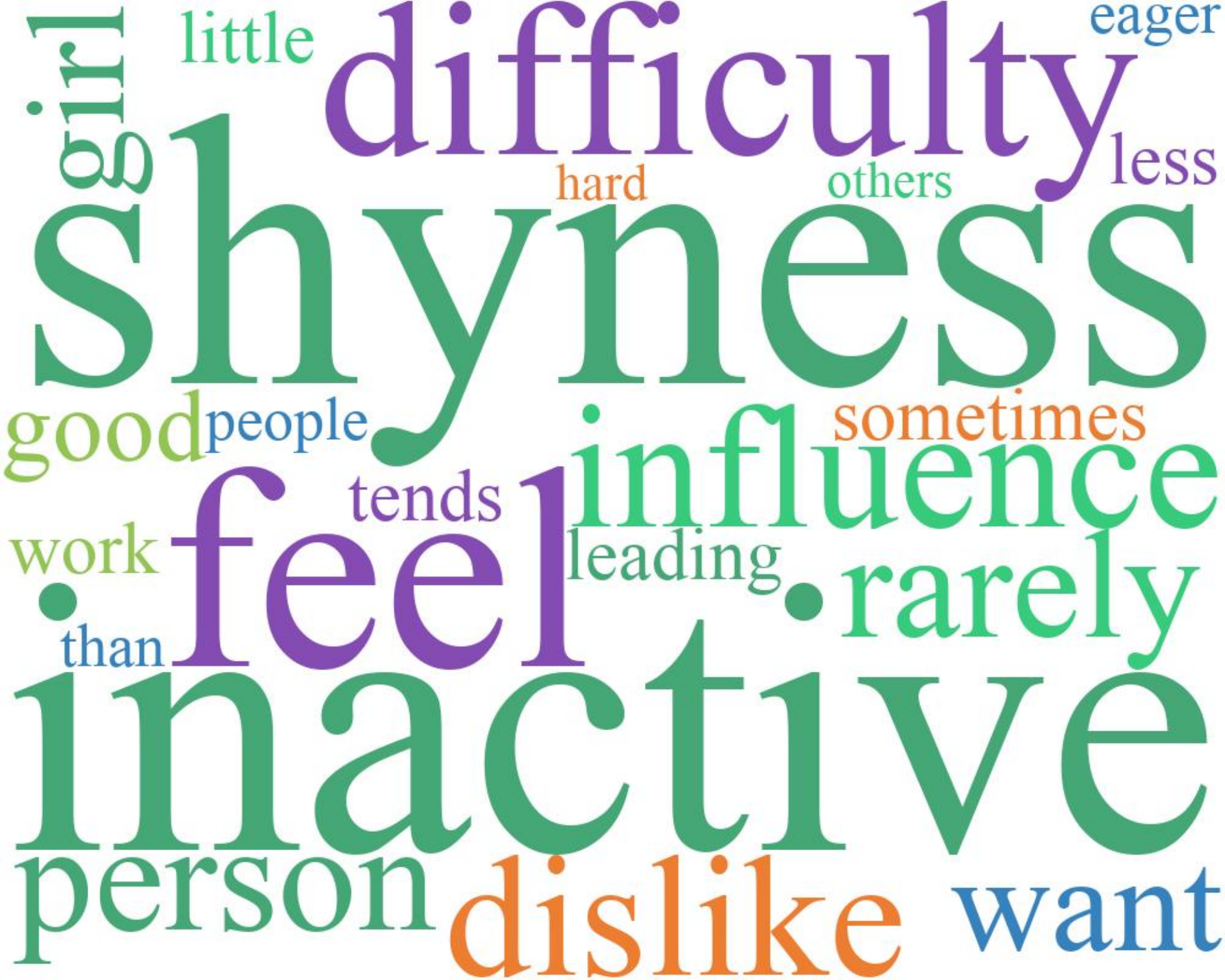}} \hspace{5pt}
\subfloat[High Agre.]{\includegraphics[width=0.15\linewidth]{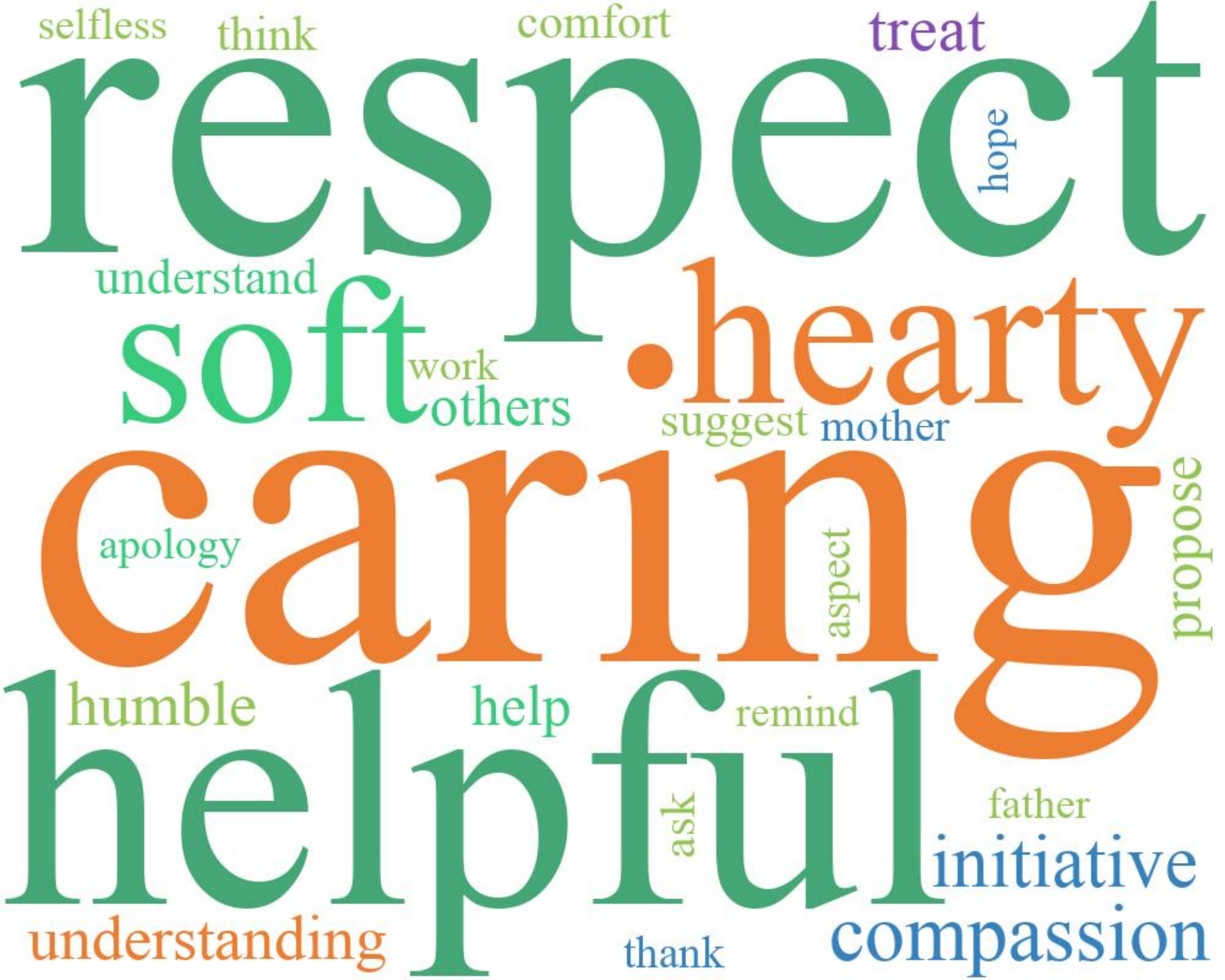}} \hspace{5pt}
\subfloat[Low Agre.]{\includegraphics[width=0.15\linewidth]{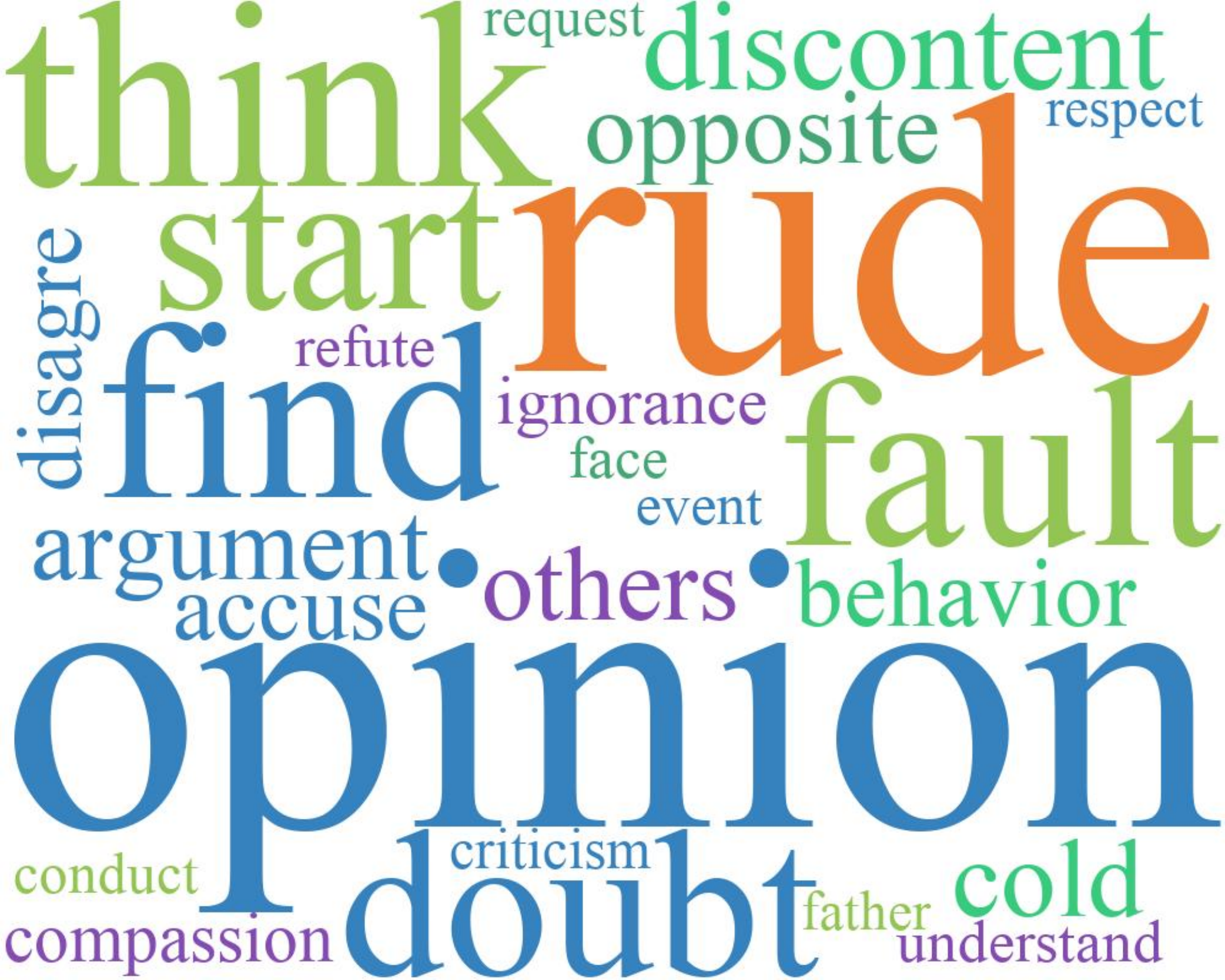}}
\hspace{5pt}
\subfloat[High Neu.]{\includegraphics[width=0.15\linewidth]{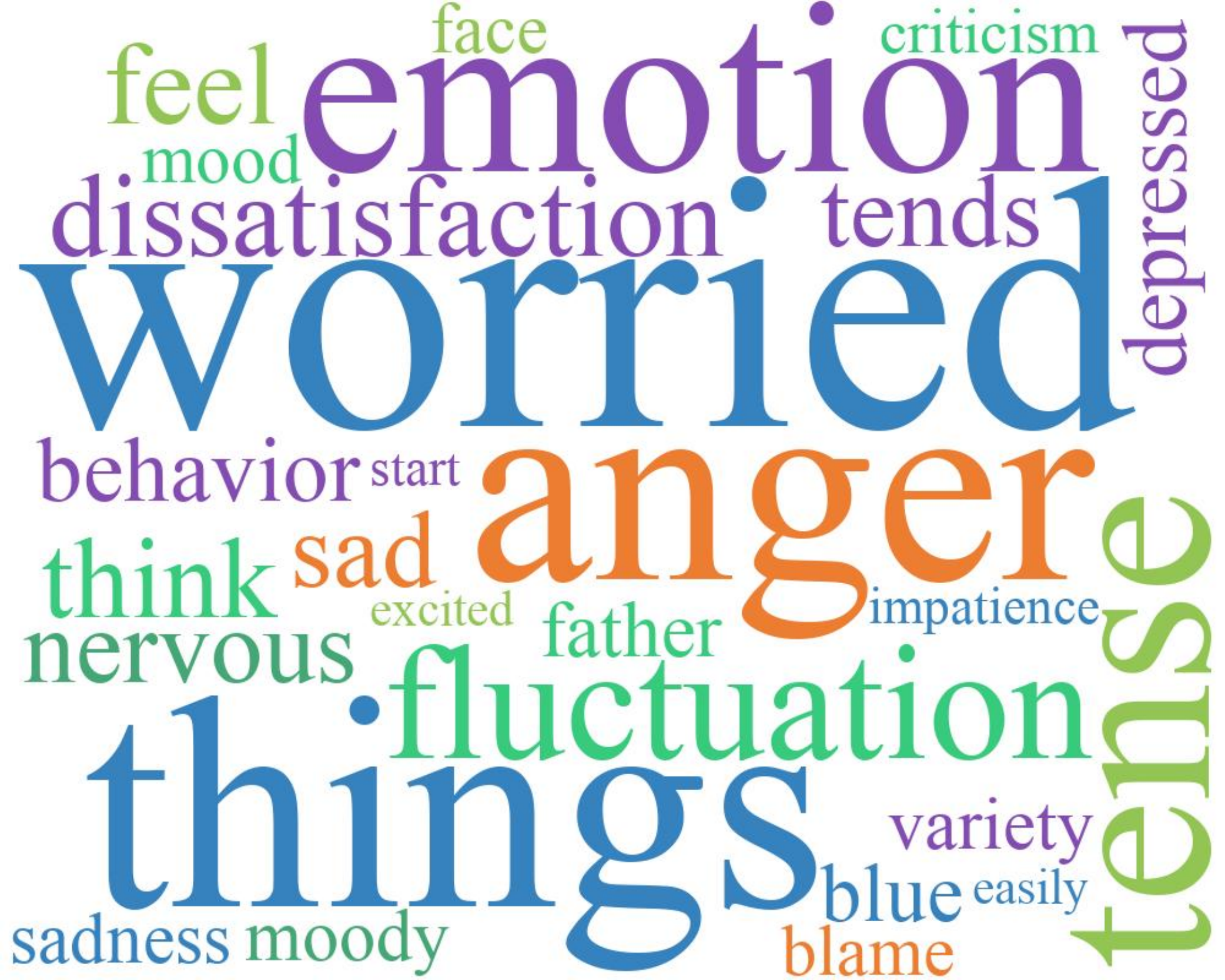}} \hspace{5pt}
\subfloat[Low Neu.]{\includegraphics[width=0.15\linewidth]{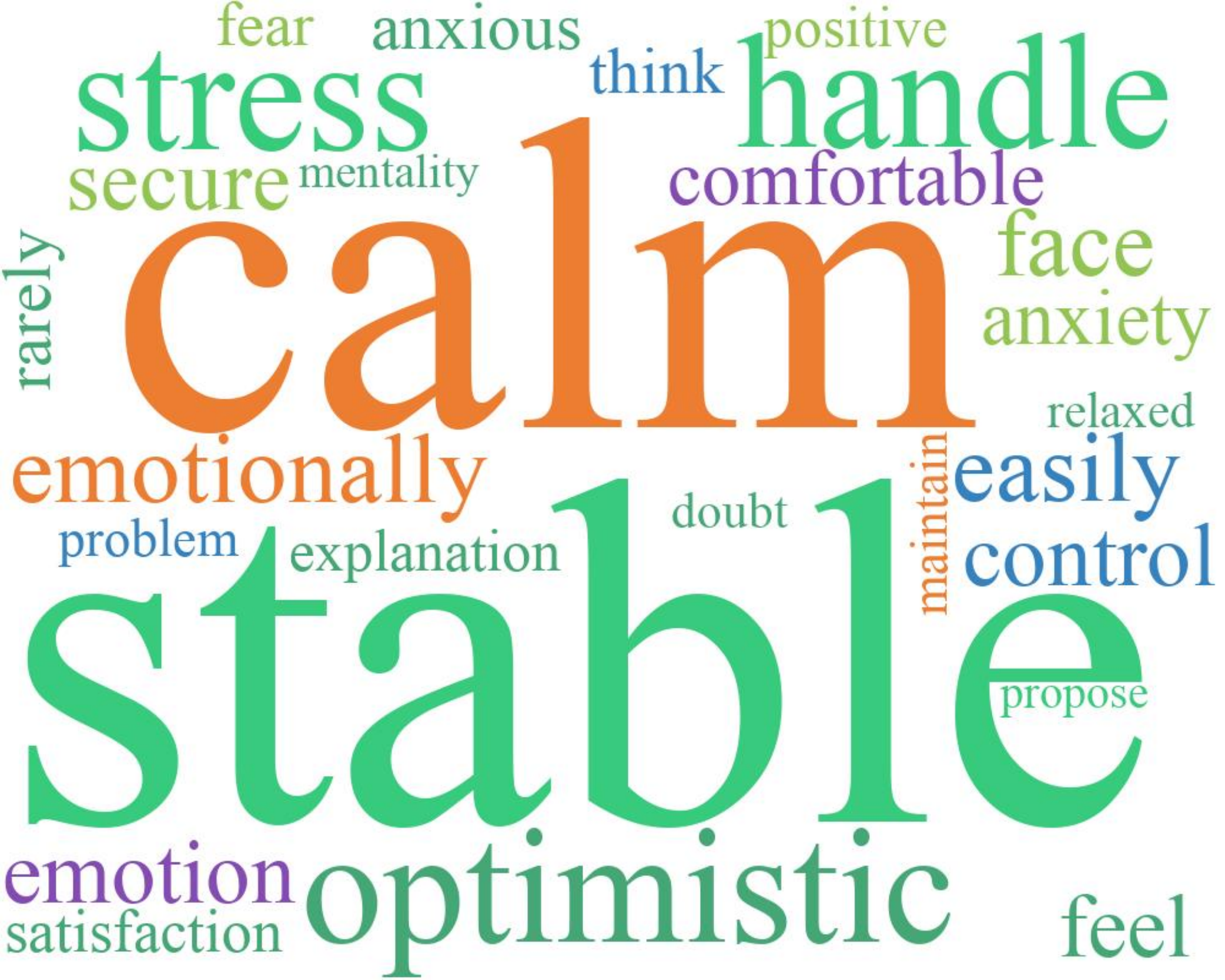}}
\caption{The word clouds of personality state of the remaining three Big-Five dimensions. (Ext.: Extraversion, Agr.: Agreeableness, and Neu.: Neuroticism)}
\label{fig:wordcloud_left}
\end{figure*}

We show the word clouds of the remaining three Big-Five dimensions: extraversion, agreeableness, and neuroticism in Figure~\ref{fig:wordcloud_left}. We filtered out stop words to highlight the key points of the natural language reasoning process.

\section{Implementation Details}
For both tasks, we use the LoRA~\cite{hu2021lora} method to fine-tune GLM-32k and Qwen-32k parameters efficiently. LoRA fine-tuning is a popular method to reduce the dependency on many training resources and yield high-quality results. We use 2*A6000 GPUs for our experiments. For the EPR-S task, the model is trained for 15 epochs with the learning rate set as 1e-4 and the batch size as 48. For the EPR-T task, we use the same training epoch and learning rate. Due to the long context of around 30 dialogues, we set the batch size as 8. To obtain more stable predictions, we make the hyperparameter
\textit{do\_sample} false to use greedy decoding at inference time.

\section{Case study}
\begin{figure*}[t]
    \centering
    \includegraphics[width=1\linewidth]{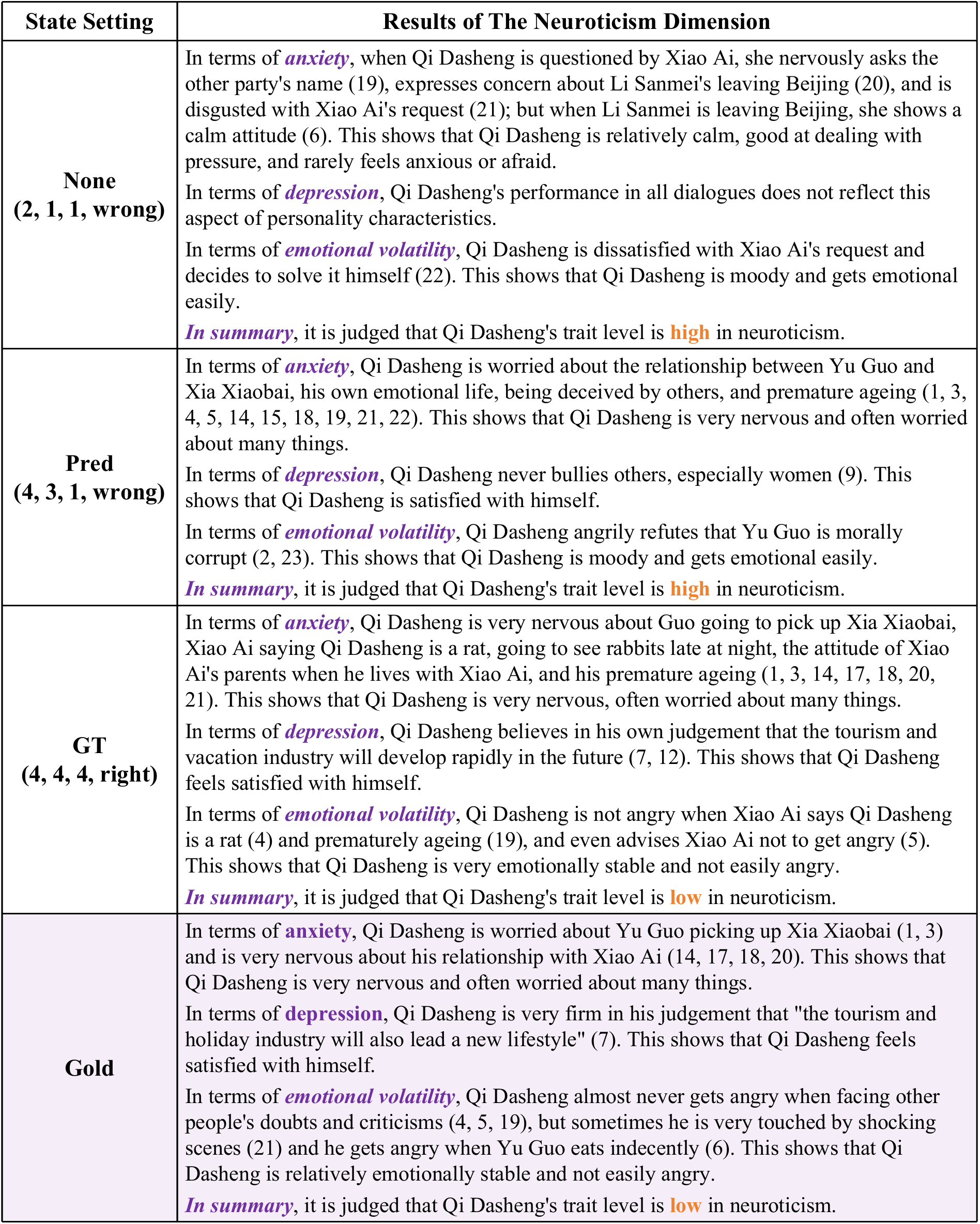}
    \caption{A case of the ablation study on the potential benefits of the state evidence for the EPR-T task. The values under each setting (score1, score2, score3, correctness) refer to the GPT-4-Turbo evaluation scores of each facet and whether the trait label is correctly predicted.
    None: no state clues, just dialogues; Pred: predicted state evidence from GLM-32k; GT: ground-truth state evidence.}
    \label{fig:case-study}
\end{figure*}

Figure \ref{fig:case-study} presents a case of the ablation study. In the None setting, ChatGLM3-6B-32K is almost unable to directly analyze useful personality clues from multiple dialogues, and these clues even contain contradictory content. Comparing the results under the Pred setting and the GT setting, we find that the ChatGLM3-6B-32K is relatively good at inferring trait evidence based on state evidence, but the model still faces great challenges in analyzing state evidence based on one dialogue.

\end{document}